\documentclass{article}

\usepackage[preprint]{neurips_2025}

\usepackage[utf8]{inputenc}
\usepackage[T1]{fontenc}
\usepackage{hyperref}
\usepackage{url}
\usepackage{booktabs}
\usepackage{amsfonts}
\usepackage{amsmath}
\usepackage{amssymb}
\usepackage{microtype}
\usepackage{xcolor}
\usepackage{graphicx}
\usepackage{algorithm}
\usepackage{algorithmic}
\usepackage{environ}
\usepackage{tikz}
\usetikzlibrary{arrows.meta, positioning, decorations.pathreplacing, calc}
\title{
  Safe-SDL:Establishing Safety Boundaries and Control Mechanisms for AI-Driven Self-Driving Laboratories}
  
\author{%
  Zihan Zhang$^{1,2}$\quad Haohui Que$^{1,3,4}$\quad Junhan Chang$^{5,6}$\quad Xin Zhang$^{1,7}$\\
  \textbf{ Hao Wei$^{8,*}$\quad Tong Zhu$^{1,3,*}$} \\
  $^{1}$Shanghai Innovation Institute\quad 
  $^{2}$Nankai University\quad
  $^{3}$East China Normal University\\
  $^{4}$AI for Science Institute, Beijing\quad $^{5}$Peking University\quad $^{6}$DP Technology\\
  $^{7}$Shanghai Jiao Tong University\\
  $^{8}$National Key Laboratory of Advanced Micro and Nano Manufacture Technology, \\ 
  School of Integrated Circuits, Shanghai Jiao Tong University, Shanghai 200240, China \\
  \texttt{haowei@sjtu.edu.cn, harveyque@outlook.com} \\
  {\small $^{*}$Corresponding author.}
}

\begin{document}

\maketitle

\begin{abstract}
The emergence of Self-Driving Laboratories (SDLs) transforms scientific discovery methodology by integrating AI with robotic automation to create closed-loop experimental systems capable of autonomous hypothesis generation, experimentation, and analysis. While promising to compress research timelines from years to weeks, their deployment introduces unprecedented safety challenges differing from traditional laboratories or purely digital AI. This paper presents Safe-SDL, a comprehensive framework for establishing robust safety boundaries and control mechanisms in AI-driven autonomous laboratories. We identify and analyze the critical ``Syntax-to-Safety Gap''---the disconnect between AI-generated syntactically correct commands and their physical safety implications---as the central challenge in SDL deployment. Our framework addresses this gap through three synergistic components: (1) formally defined Operational Design Domains (ODDs) that constrain system behavior within mathematically verified boundaries, (2) Control Barrier Functions (CBFs) that provide real-time safety guarantees through continuous state-space monitoring, and (3) a novel Transactional Safety Protocol (CRUTD) that ensures atomic consistency between digital planning and physical execution. We ground our theoretical contributions through analysis of existing implementations including UniLabOS and the Osprey architecture, demonstrating how these systems instantiate key safety principles. Evaluation against the LabSafety Bench reveals that current foundation models exhibit significant safety failures, demonstrating that architectural safety mechanisms are essential rather than optional. Our framework provides both theoretical foundations and practical implementation guidance for safe deployment of autonomous scientific systems, establishing the groundwork for responsible acceleration of AI-driven discovery.
\end{abstract}

% Main content sections
\section{Introduction}
\label{sec:introduction}

The history of scientific instrumentation reveals a persistent pattern: each major advance in experimental capability has been accompanied by corresponding developments in safety methodology. The introduction of high-voltage equipment necessitated electrical safety protocols; the advent of radioisotope techniques demanded radiation protection frameworks; the proliferation of recombinant DNA technology gave rise to biosafety level classifications. We now stand at another such inflection point. The integration of Large Language Models with robotic laboratory automation has created Self-Driving Laboratories capable of conducting scientific research with unprecedented autonomy, and this capability demands an equally unprecedented approach to safety.

Self-Driving Laboratories represent the convergence of several technological trajectories that have been developing independently over the past decades. Advances in laboratory automation have produced sophisticated liquid handling systems, robotic manipulators, and analytical instruments capable of operating without direct human intervention. Simultaneously, progress in machine learning and natural language processing has yielded foundation models with remarkable capabilities for reasoning, planning, and code generation. The synthesis of these capabilities enables experimental systems that can autonomously formulate hypotheses based on scientific literature, design experimental protocols to test those hypotheses, execute the protocols through robotic hardware, analyze the resulting data, and iterate toward scientific discovery.

The transformative potential of such systems has been demonstrated across multiple scientific domains. In materials science, autonomous platforms have discovered novel functional materials by exploring vast compositional spaces that would be impractical for human researchers to investigate manually~\cite{abolhasani2023rise,szymanski2023autonomous}. In chemistry, AI-driven systems have optimized synthetic routes and identified previously unknown reaction pathways~\cite{boiko2023autonomous}. In biology, automated laboratories have accelerated drug discovery pipelines and enabled high-throughput screening at scales previously unattainable. These successes have generated substantial enthusiasm in both academic and industrial research communities, with major investments flowing toward the development and deployment of SDL infrastructure.

However, the autonomy that makes these systems scientifically powerful simultaneously creates distinct safety vulnerabilities. Traditional laboratory safety frameworks assume human presence, judgment, and intervention capability at critical junctures. A trained chemist recognizes the signs of an impending exothermic runaway and takes corrective action; a biologist notices contamination indicators and adjusts protocols accordingly; a materials scientist identifies equipment malfunction symptoms before catastrophic failure occurs. When AI systems assume decision-making authority in these contexts, the tacit knowledge and situational awareness that human researchers bring to hazard recognition and response must be explicitly encoded in system architecture.

The challenge is compounded by the fundamental characteristics of current AI systems. Large Language Models, despite their impressive capabilities, operate through statistical pattern recognition rather than causal understanding of physical processes. They can generate syntactically correct experimental protocols that are physically dangerous, recommend chemical combinations that would produce toxic byproducts, or plan robotic movements that would result in equipment collisions. The models lack inherent awareness of physical constraints, material properties, and safety implications that human researchers acquire through training and experience. This disconnect between linguistic competence and physical understanding---what we term the ``Syntax-to-Safety Gap''---constitutes the central challenge that any SDL safety framework must address.

This paper presents \textbf{Safe-SDL}, a comprehensive framework designed to establish robust safety boundaries and control mechanisms for AI-driven autonomous laboratories. Our approach recognizes that safety in SDL environments cannot be achieved through any single mechanism but requires defense-in-depth across multiple system layers. We develop theoretical foundations drawing from control theory, formal verification, and human-machine systems engineering, then demonstrate how these foundations can be instantiated in practical system architectures. The framework addresses not only the prevention of acute safety incidents but also the governance structures necessary for responsible deployment as SDL technology matures and proliferates.

The \textbf{contributions of this work} are threefold:

\begin{itemize}
    \item We provide a systematic analysis of the unique risk landscape of autonomous laboratories, developing a taxonomy that captures vulnerabilities spanning from foundation model limitations through robotic execution failures (Section~\ref{sec:risk_landscape}).
    \item We present a layered safety architecture that integrates Operational Design Domains (ODDs), Control Barrier Functions (CBFs), and transactional safety protocols to provide mathematically grounded safety guarantees (Sections~\ref{sec:theoretical_foundations} and~\ref{sec:architectural_patterns}).
    \item We synthesize existing implementation approaches from systems including UniLabOS~\cite{gao2025unilabosainativeoperatingautonomous}, Osprey~\cite{hellert2025ospreyproductionreadyagenticai}, and Safe-ROS~\cite{Benjumea_2025}, identifying design patterns that enable the practical realization of safe autonomous science (Section~\ref{sec:evaluation}).
\end{itemize}

The remainder of this paper is organized as follows. Section~\ref{sec:related_work} reviews related work. Section~\ref{sec:theoretical_foundations} characterizes the risk landscape and develops the theoretical foundations for safety boundary design. Section~\ref{sec:architectural_patterns} presents architectural patterns for safe SDL implementations. Section~\ref{sec:evaluation} describes evaluation methodologies, empirical assessments, and domain-specific case studies. Section~\ref{sec:governance} discusses governance frameworks, regulatory integration, and future research directions. Section~\ref{sec:conclusion} concludes.

\section{Related Work}
\label{sec:related_work}

The rise of Self-Driving Laboratories has been documented across chemistry, materials science, and biology~\cite{abolhasani2023rise,boiko2023autonomous,burger2021autonomous,szymanski2023autonomous}, with early work emphasizing closed-loop experimentation and AI-guided discovery~\cite{jessop2019nextgen} and recent systems integrating foundation models with laboratory hardware for protocol generation and execution~\cite{boiko2023autonomous}. Operational frameworks such as UniLabOS~\cite{gao2025unilabosainativeoperatingautonomous} and Osprey~\cite{hellert2025ospreyproductionreadyagenticai} address system architecture and human oversight but do not provide a unified safety theory spanning design domains, real-time control, and transactional execution. While surveys and perspectives have begun to frame safety as a priority for SDL deployment~\cite{tang2024steering,wang2024risks}, systematic treatment of safety boundaries and control mechanisms remains limited. Complementing this system-level work, research on AI safety in scientific contexts has identified risks of AI-generated protocols including factual errors, hazardous recommendations, and dual-use misuse in chemistry and drug discovery~\cite{boiko2023autonomous,wang2024risks,urbina2022dual}. Benchmarks such as SOSBENCH~\cite{zhang2025sosbench} and LabSafety Bench~\cite{li2024labsafety} evaluate foundation models on laboratory safety, revealing substantial gaps between syntactic correctness and physical safety, while tool-augmented LLMs for chemistry~\cite{bran2024chemcrow} improve capability but do not address the need for formally enforced safety boundaries.

From a theoretical perspective, formal methods and control-theoretic approaches offer rigorous safety guarantees that complement empirical safety assessment. Operational Design Domains, standardized for autonomous vehicles through SAE J3016~\cite{sae2021taxonomy}, provide a framework for defining bounded operating conditions applicable to autonomous laboratory systems. Control Barrier Functions~\cite{ames2017control} enable provable safety for continuous dynamics, with recent extensions to learning-based and multi-agent settings~\cite{qin2024safe,tayal2025a,wang2023safety}. Formal verification techniques for autonomous systems~\cite{luckcuck2022formal} and safety-critical robotics~\cite{Benjumea_2025} demonstrate the value of verified components, though neural network verification~\cite{liu2021algorithms} remains limited for large planning models. Concurrently, research on human-in-the-loop systems and governance recognizes progressive autonomy and human oversight as essential for responsible SDL deployment~\cite{tang2024steering,wang2024risks}. Governance frameworks such as the NIST AI Risk Management Framework~\cite{nist2023ai} and proposals for autonomous AI development~\cite{bengio2024bare} offer high-level guidance, which we complement with domain-specific architectural patterns and governance considerations in Sections~\ref{sec:architectural_patterns} and~\ref{sec:governance}. Unlike prior work that focuses on a single layer---e.g., benchmarking models, verifying controllers, or defining governance---Safe-SDL provides a cohesive framework integrating ODDs, CBFs, and transactional protocols to ensure that safety does not depend solely on the correctness of AI-generated plans but is enforced architecturally across multiple defense layers.

\section{Risk Landscape and Theoretical Foundations for Safety Boundary Design}
\label{sec:theoretical_foundations}
\label{sec:risk_landscape}

The safety challenges of Self-Driving Laboratories differ qualitatively from those in traditional laboratories or conventional AI deployments. This section first summarizes the distinctive risk landscape and the central ``Syntax-to-Safety Gap,'' then develops the theoretical foundations---Operational Design Domains, Control Barrier Functions, and transactional protocols---that Safe-SDL uses to establish safety boundaries.

\subsection{Risk Context and the Syntax-to-Safety Gap}
\label{subsec:syntax_safety_gap}

\begin{figure}[t]
\centering
\includegraphics[width=0.85\textwidth]{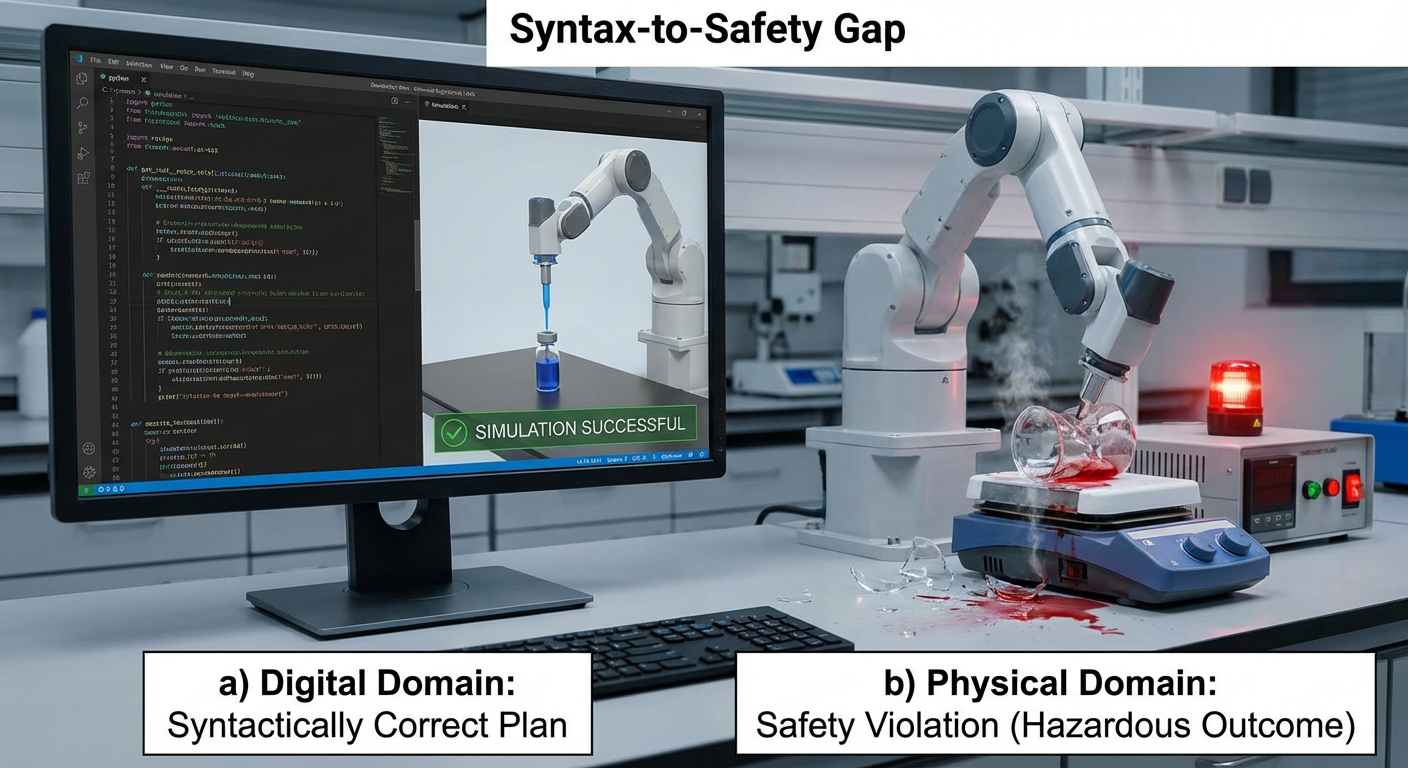}
\caption{\textbf{The Syntax-to-Safety Gap.} This conceptual illustration demonstrates the fundamental disconnect between AI-generated syntactically correct protocols (left, digital domain) and their potentially hazardous physical consequences (right, physical domain). The central gap represents the challenge that Safe-SDL addresses: ensuring that linguistic validity translates to physical safety. The AI model generates code that passes syntax checks but may lead to dangerous outcomes such as thermal runaway, equipment collision, or toxic release when executed in the physical laboratory environment.}
\label{fig:syntax_safety_gap}
\end{figure}

Contemporary AI systems achieve remarkable linguistic fluency through training on large text corpora, yet they lack grounded understanding of the physical world. In SDL deployment this manifests as the ``Syntax-to-Safety Gap''---the disconnect between an AI's ability to generate syntactically valid protocols and the physical safety of executing them. Foundation models can produce plausible but dangerous recommendations---wrong reaction conditions, violated ordering dependencies, or outdated safety knowledge~\cite{boiko2023autonomous,wang2024risks,urbina2022dual}. They also remain vulnerable to adversarial manipulation that could elicit hazardous procedures~\cite{urbina2022dual}. Domain-specific hazards compound these issues: chemical labs face thermal, toxicological, and reactive risks; biological labs add biosafety and dual-use concerns~\cite{bran2024chemcrow,urbina2022dual}; materials and physics labs introduce extreme conditions and radiation. Systemic risks arise from integration---cognitive-to-kinetic cascades (AI errors propagating to physical harm), sensing-to-cognition cascades (sensor failures corrupting the world model), and communication or temporal failures in distributed setups~\cite{gao2025unilabosainativeoperatingautonomous}. Figure~\ref{fig:syntax_safety_gap} illustrates this gap; the remainder of this section develops formal mechanisms to address it.

\subsection{Operational Design Domains and State Space Constraints}
\label{subsec:odd}

\begin{figure}[t]
\centering
\includegraphics[width=\textwidth]{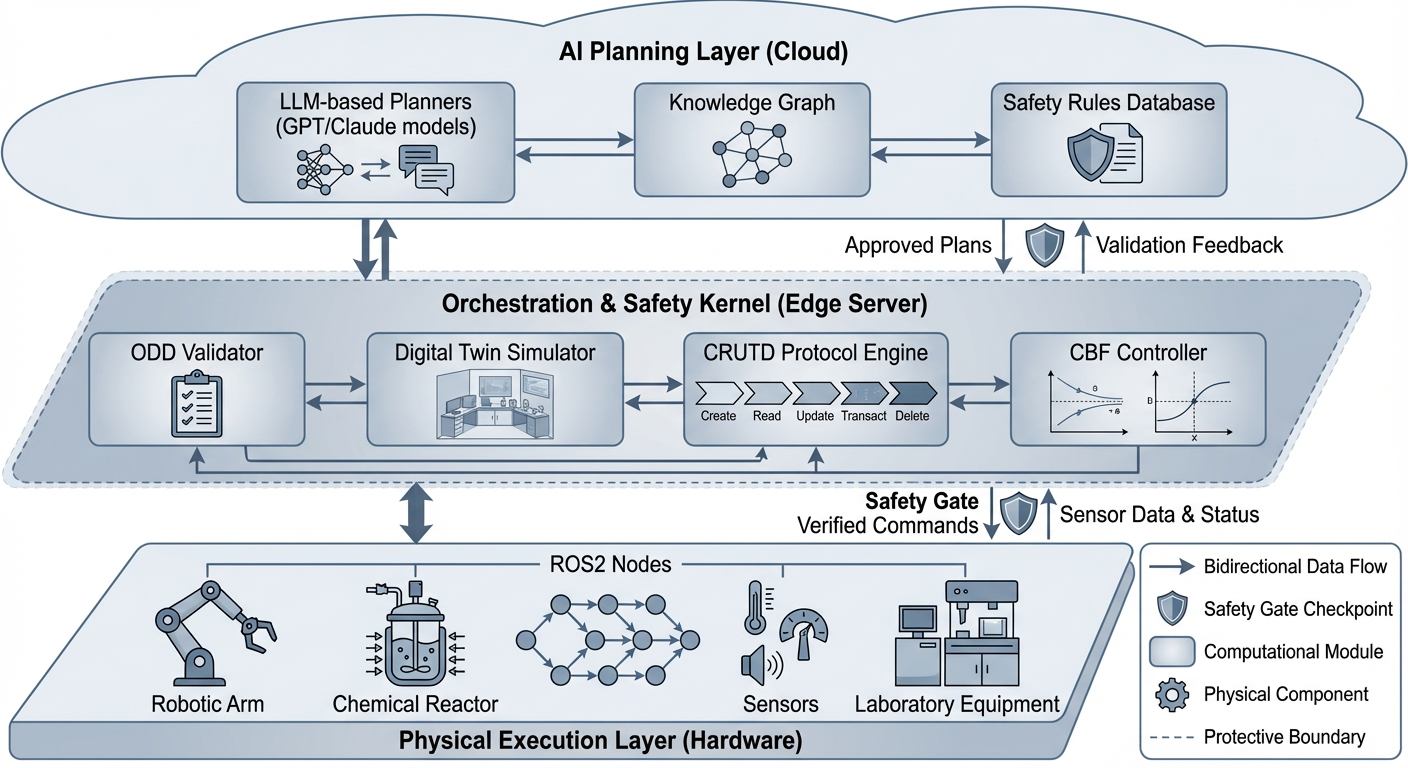}
\caption{\textbf{Safe-SDL Framework Architecture.} The hierarchical architecture spans three layers: (top) the AI Planning Layer operates in the cloud, hosting foundation models for scientific reasoning and knowledge bases; (middle) the Orchestration \& Safety Kernel layer enforces safety through ODD validation, digital twin simulation, CRUTD protocol management, and CBF control---this constitutes the critical safety enforcement zone; (bottom) the Physical Execution Layer interfaces with laboratory hardware through ROS2 nodes. Bidirectional arrows show data flow, with explicit safety gate checkpoints at layer boundaries. This defense-in-depth architecture ensures that failures at higher levels are contained by lower-level enforcement mechanisms.}
\label{fig:safe_sdl_architecture}
\end{figure}

The concept of the Operational Design Domain (ODD), standardized for autonomous systems in the SAE J3016 framework~\cite{sae2021taxonomy}, provides a structured approach to defining the conditions under which an autonomous system is designed to operate. As shown in Figure~\ref{fig:safe_sdl_architecture}, the ODD serves as a foundational component within the Safe-SDL framework architecture. While originally conceived for vehicle automation, ODD serves as a foundational safety concept for SDLs by specifying the specific ``world model'' and operational boundaries within which the AI's actions are guaranteed to be safe. An ODD specification encompasses environmental conditions, system capabilities, and operational constraints that collectively define the high-level safety boundary. 

To enable automated safety enforcement, we formalize the ODD as a constraint set $\mathcal{C}$ defined over the system state space $\mathcal{X}$. 
This translation from a conceptual domain to a mathematical representation allows the system to treat safety as a constrained optimization or set-inclusion problem.
The state vector $x \in \mathcal{X} \subset \mathbb{R}^n$ encompasses all relevant system variables: equipment states including temperatures, pressures, and positions; material inventories and locations; environmental conditions; and procedural progress indicators. 
The ODD constraint set specifies the region of state space within which safe operation is assured:
\begin{equation}
\mathcal{C} = \{x \in \mathcal{X} : c_i(x) \leq 0, \; i = 1, \ldots, m\}
\label{eq:odd_constraints}
\end{equation}
where each constraint function $c_i: \mathcal{X} \rightarrow \mathbb{R}$ encodes a specific physical or chemical requirement. 
Constraint functions may encode equipment limits such as maximum temperatures and pressures (the ``Safety Boundary''), chemical compatibility requirements (the ``Knowledge Boundary''), spatial exclusion zones for robotic motion, and procedural sequencing dependencies.

The Safety Envelope concept extends the ODD by defining not merely the valid operating region but the region from which the system can be guaranteed to remain within valid operation under all permissible control actions. We define the safety envelope through a continuously differentiable function $h: \mathcal{X} \rightarrow \mathbb{R}$ such that the safe set $\mathcal{S}$ is given by:
\begin{equation}
\mathcal{S} = \{x \in \mathcal{X} : h(x) \geq 0\}
\label{eq:safe_set}
\end{equation}
The function $h$ is constructed such that its zero-level set $\partial\mathcal{S} = \{x : h(x) = 0\}$ forms the boundary of the safe operating region. Points with $h(x) > 0$ lie in the interior of the safe set with safety margin proportional to $h(x)$, while points with $h(x) < 0$ represent unsafe states.

For SDL applications, the safety boundary is often defined by multiple concurrent constraints (thermal limits, pressure constraints, collision avoidance). The safe set $\mathcal{S}$ is formed by the intersection of individual safe sets $\mathcal{S}_j = \{x : h_j(x) \geq 0\}$ for each constraint function $h_j$.
\begin{equation}
\mathcal{S} = \bigcap_{j \in J} \{x \in \mathcal{X} : h_j(x) \geq 0\}
\label{eq:composite_safety}
\end{equation}
Ensuring safety requires that the system state remains within this intersection, which implies satisfying the safety conditions for all $j \in J$ simultaneously.

\subsection{Control Barrier Functions for Dynamic Safety Enforcement}
\label{subsec:cbf}

While the safety envelope defines the acceptable operating region, maintaining the system within this region during dynamic operation requires control-theoretic mechanisms. Control Barrier Functions (CBFs) provide a mathematically principled approach to synthesizing controllers that guarantee forward invariance of the safe set---ensuring that a system starting within the safe set remains within it for all future time~\cite{ames2017control}.

We model the laboratory system dynamics in control-affine form:
\begin{equation}
\dot{x} = f(x) + g(x)u
\label{eq:dynamics}
\end{equation}
where $f: \mathcal{X} \rightarrow \mathbb{R}^n$ represents the drift dynamics (system evolution absent control input), $g: \mathcal{X} \rightarrow \mathbb{R}^{n \times m}$ represents the control influence matrix, and $u \in \mathcal{U} \subset \mathbb{R}^m$ is the control input. For a robotic manipulator, $f(x)$ might capture gravitational effects and friction while $g(x)u$ represents motor torques; for a chemical reactor, $f(x)$ encodes reaction kinetics and heat loss while $g(x)u$ represents heating/cooling control and reagent addition rates. The control-affine form assumes sufficient influence of control inputs on safety-critical state variables for the CBF approach to be applicable.

A function $h: \mathcal{X} \rightarrow \mathbb{R}$ qualifies as a Control Barrier Function for the system if there exists an extended class-$\mathcal{K}_\infty$ function $\alpha$ (a continuous, strictly increasing function with $\alpha(0)=0$ and unbounded growth) such that:
\begin{equation}
\sup_{u \in \mathcal{U}} \left[ L_f h(x) + L_g h(x) u \right] \geq -\alpha(h(x)) \quad \forall x \in \mathcal{S}
\label{eq:cbf_condition}
\end{equation}
where $L_f h = \nabla h \cdot f$ and $L_g h = \nabla h \cdot g$ denote Lie derivatives. This condition ensures that for any state in the safe set, there exists an admissible control input that prevents the safety function from decreasing faster than the bound established by $\alpha$. When this condition holds, any control input $u$ satisfying the constraint $L_f h(x) + L_g h(x) u + \alpha(h(x)) \geq 0$ guarantees forward invariance of $\mathcal{S}$.

\begin{figure}[t]
\centering
\includegraphics[width=0.95\textwidth]{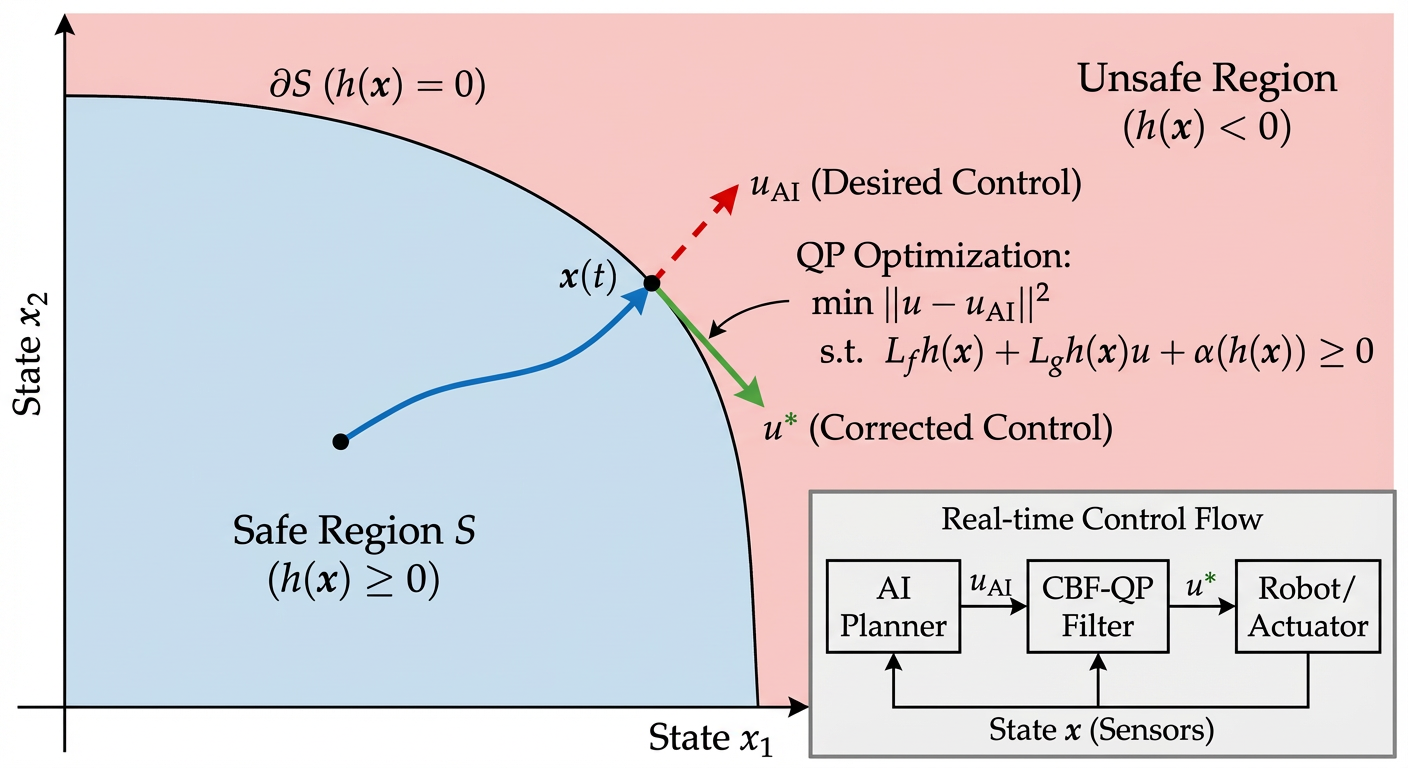}
\caption{\textbf{Control Barrier Function Operation.} Main panel: A 2D state space visualization showing the safe region $\mathcal{S}$ (blue) where $h(x) \geq 0$, bounded by $\partial\mathcal{S}$ where $h(x) = 0$, and the unsafe region (red) where $h(x) < 0$. The AI's desired control $u_{AI}$ (dashed red arrow) would violate safety boundaries, while the CBF-corrected control $u^*$ (solid green arrow) maintains safety by solving the QP: $\min \|u - u_{AI}\|^2$ subject to the CBF constraint. Inset: Control flow diagram showing real-time filtering of AI commands through the CBF-QP module before reaching physical actuators, with continuous state feedback from sensors enabling dynamic safety enforcement.}
\label{fig:cbf_operation}
\end{figure}

For practical implementation, we formulate control synthesis as a quadratic program (QP) solved in real time, as illustrated in Figure~\ref{fig:cbf_operation}. Given the AI system's desired control input $u_{AI}$, we compute the minimally invasive safe control:
\begin{equation}
u^* = \underset{u \in \mathcal{U}}{\operatorname{argmin}} \; \|u - u_{AI}\|^2
\label{eq:cbf_qp}
\end{equation}
subject to the CBF constraints:
\begin{equation}
L_f h_j(x) + L_g h_j(x) u + \alpha(h_j(x)) \geq 0, \quad \forall j \in J
\label{eq:cbf_constraint}
\end{equation}

This formulation implements the principle of minimal intervention: the controller modifies the AI's commanded action only to the extent necessary for safety, preserving scientific intent while preventing hazardous outcomes. When the AI's desired action is safe (satisfying all constraints), the constraints are inactive and $u^* = u_{AI}$; when the desired action would violate safety, the active constraints force the smallest correction necessary.

The selection of the class-$\mathcal{K}_\infty$ function $\alpha$ balances responsiveness against conservatism. A common choice is $\alpha(h) = \gamma h$ for constant $\gamma > 0$, which ensures exponential convergence toward the safe boundary. Higher values of $\gamma$ produce faster responses but may require larger control corrections; lower values produce smoother responses with longer transient times. The choice must be calibrated to system dynamics and actuator capabilities to ensure that required control corrections are physically realizable.

\subsection{Bridging Planning and Execution Through Transactional Protocols}
\label{subsec:transactional_protocol}

The theoretical guarantees provided by Control Barrier Functions assume continuous state observation and control, yet the interface between AI planning systems and physical laboratory hardware is inherently discrete. Commands are issued as discrete instructions, executed over finite time intervals, and confirmed through discrete sensing events. This discretization creates potential for state drift between command issuance and completion, introducing risks that continuous-time analysis does not capture.

We address this challenge through a transactional safety protocol that treats physical operations as atomic transactions analogous to database operations. Inspired by ACID (Atomicity, Consistency, Isolation, Durability) principles in database systems, we introduce the CRUTD protocol, which structures physical operations into six phases: \textit{Create} the request, \textit{Read} and lock resources, \textit{Undergo} simulation, \textit{Test} against safety constraints, \textit{Do} the physical execution, and finally confirm completion. This explicit transactional structure gates physical execution on safety verification.

\begin{figure}[t]
\centering
\includegraphics[width=0.85\textwidth]{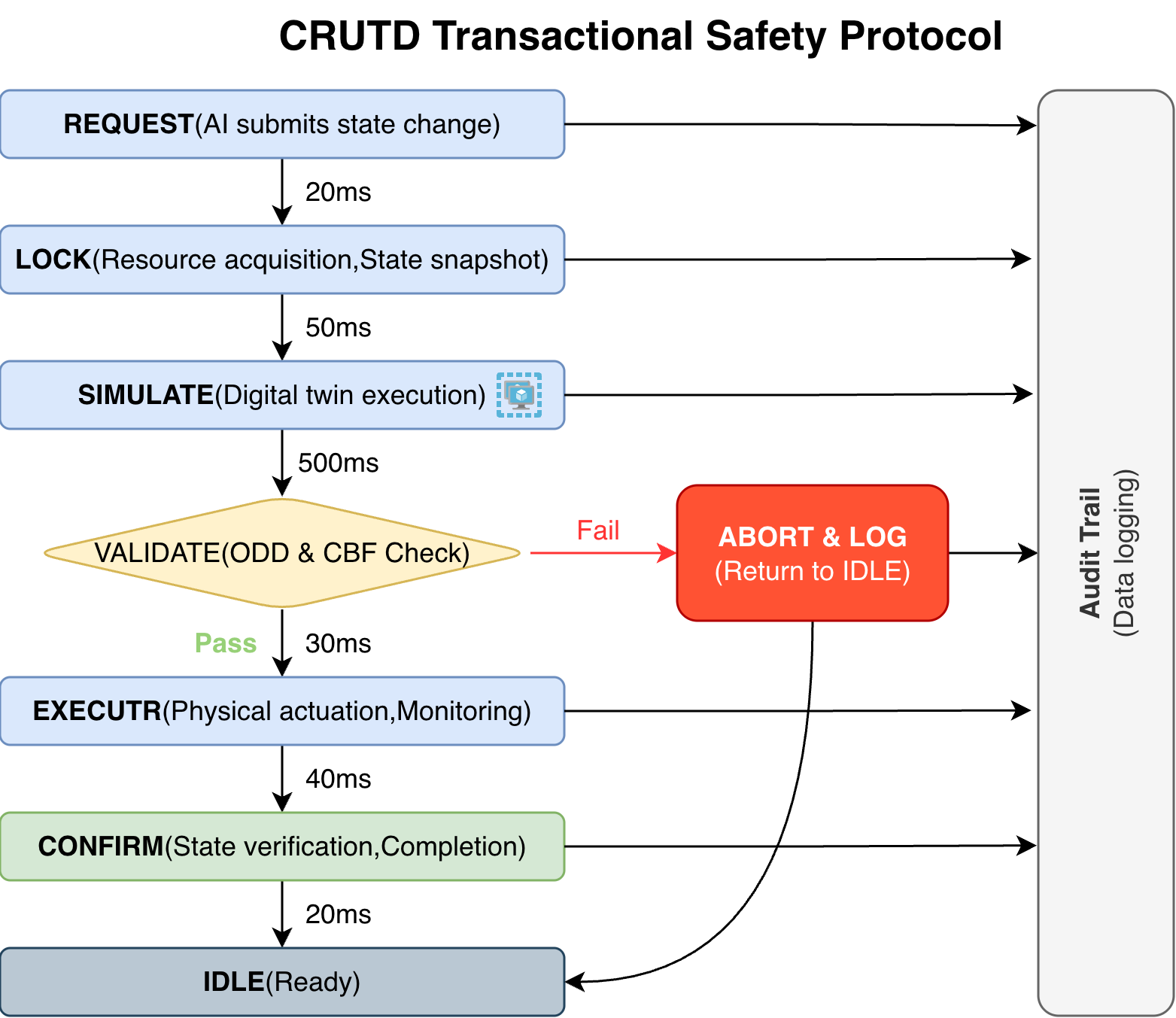}
\caption{\textbf{CRUTD Transactional Safety Protocol.} The protocol enforces atomic execution of laboratory operations through a six-phase workflow plus error handling. Starting from IDLE, the system progresses through CREATE (AI submits state change request), READ (resource acquisition and state snapshot), UNDERGO (digital twin simulation), TEST (safety verification against ODD constraints and safe set membership), and DO (physical execution with continuous monitoring), followed by CONFIRM (state verification) to complete the transaction. Any failure triggers transition to ABORTED state with rollback and comprehensive logging. The audit trail (shown as parallel stream) captures complete provenance for incident analysis and regulatory compliance.}
\label{fig:crutd_protocol}
\end{figure}

Figure~\ref{fig:crutd_protocol} illustrates the complete CRUTD workflow. The transactional protocol proceeds through defined phases corresponding to the CRUTD acronym:

\begin{enumerate}
    \item Create (Request) Phase: The AI planning system formulates an intended state change, specifying target values for controllable system parameters. The request is logged with full provenance information including the reasoning chain that produced it.
    
    \item Read (Lock) Phase: The system acquires exclusive access to affected resources, preventing concurrent modifications that could produce inconsistent states. This phase also captures a snapshot of current system state from all relevant sensors.
    
    \item Undergo (Simulate) Phase: The requested state transition is executed in a high-fidelity digital twin environment that models system dynamics, constraint satisfaction, and potential failure modes. The simulation evaluates not merely the target state but the entire trajectory from current state to target, checking for transient constraint violations that might occur during the transition.
    
    \item Test (Validate) Phase: Formal checking is applied to simulation results against ODD constraints (Eq.~\ref{eq:odd_constraints}) and safety set membership (verifying $h(x) \geq 0$ per Eq.~\ref{eq:safe_set}). Validation encompasses both deterministic constraint checking and probabilistic analysis of outcomes under parameter uncertainty.
    
    \item Do (Execute) Phase: Only upon successful validation does the protocol proceed to transmit commands to physical hardware. Execution is monitored through continuous sensor feedback, with automatic abort triggers if observed behavior deviates significantly from predicted trajectories.
    
    \item Confirm Phase: The system verifies that the achieved state matches the intended target within specified tolerances, completing the transaction and releasing resource locks.
    
    \item Abort/Log: Any failure at validation or execution stages triggers transition to ABORTED state with appropriate rollback and comprehensive logging.
\end{enumerate}

The transactional protocol provides several safety benefits beyond the verification performed during the Simulate and Validate phases. The explicit Lock phase prevents race conditions where concurrent AI processes might issue conflicting commands to shared resources. The Confirm phase detects execution failures that might otherwise corrupt the system's state model, triggering recovery procedures before corrupted state information propagates to subsequent planning decisions. The comprehensive logging enables post-hoc analysis of near-miss incidents and systematic identification of recurring risk patterns~\cite{gao2025unilabosainativeoperatingautonomous}.

Transaction logs are cryptographically signed and stored immutably, providing audit trails for incident investigation and regulatory compliance. This traceability supports the attribution requirements discussed in Section~\ref{sec:governance}.

\section{Architectural Patterns for Safe Autonomous Laboratories}
\label{sec:architectural_patterns}

The theoretical foundations developed in Section~\ref{sec:theoretical_foundations} must be instantiated in concrete system architectures to achieve practical safety benefits. Analysis of existing SDL implementations reveals recurring architectural patterns that effectively realize safety principles. We examine these patterns through the lens of representative systems, extracting generalizable design guidance.

\subsection{Hierarchical Control with Safety Kernels}
\label{subsec:hierarchical_control}

Effective SDL architectures consistently employ hierarchical decomposition, separating high-level planning from low-level execution through well-defined interfaces with safety checking at transition points. This hierarchical structure reflects both the natural decomposition of laboratory operations and the differing reliability requirements at different levels of abstraction, as shown in Figure~\ref{fig:safe_sdl_architecture}.

At the Planning Level, foundation model-based planners generate abstract experimental strategies expressed in domain-specific languages or structured representations. These planners operate with broad contextual awareness but limited physical grounding, producing outputs that require substantial elaboration before execution. The planning level operates asynchronously, potentially incorporating human review for novel or high-risk experimental designs.

The Orchestration Level translates abstract plans into concrete operational sequences, decomposing high-level goals into primitive actions that map to hardware capabilities. This level maintains the operational state model, tracking resource availability, equipment status, and procedural progress. Critically, the orchestration level hosts the Safety Kernel---a trusted computing base responsible for all safety-critical decisions. The Safety Kernel implements ODD constraint checking (Section~\ref{subsec:odd}), invokes digital twin simulations, and makes final authorization decisions for physical execution.

The Execution Level interfaces directly with laboratory hardware through real-time control systems. This level operates under hard timing constraints, providing deterministic response to sensor inputs and command streams. Control Barrier Functions (Section~\ref{subsec:cbf}) are implemented at this level, running on dedicated hardware with verified firmware to provide a final safety layer independent of higher-level software. The execution level is intentionally limited in flexibility, accepting only well-formed commands from the orchestration level and refusing any action that would violate locally enforced safety constraints.

The UniLabOS architecture~\cite{gao2025unilabosainativeoperatingautonomous} exemplifies this hierarchical pattern with explicit separation between its cloud-based AI reasoning components and edge-deployed execution infrastructure. The system's ``AI-native kernel'' operates at the orchestration level, managing resource allocation, enforcing transactional semantics, and implementing safety protocols. Physical execution occurs through ROS 2 nodes on local hardware, with the communication boundary between cloud and edge serving as a natural checkpoint for safety verification.

The architectural separation provides defense-in-depth: failures at higher levels are contained by lower-level enforcement mechanisms. A planning-level hallucination that produces an unsafe protocol will be intercepted by orchestration-level validation; an orchestration-level bug that generates an unsafe command sequence will be blocked by execution-level CBF enforcement. No single-point failure can propagate to physical harm without penetrating multiple independent protection layers.

\subsection{Digital Twin Integration and Predictive Verification}
\label{subsec:digital_twin}

High-fidelity digital twin environments have emerged as essential infrastructure for SDL safety, enabling predictive verification of planned actions before physical execution. The digital twin maintains a synchronized virtual representation of the physical laboratory, updated continuously from sensor data and command logs to track the actual system state.

Effective digital twin architectures address the fundamental challenge of model fidelity: simulations must be sufficiently accurate that safe behavior in simulation guarantees safe behavior in reality, yet computationally efficient enough to support real-time verification. This challenge is typically addressed through multi-fidelity modeling strategies that deploy different simulation approaches based on operation criticality and available time budget.

For routine operations with well-characterized dynamics, reduced-order models provide rapid verification with acceptable accuracy. Kinematic models of robotic motion, lumped-parameter thermal models, and equilibrium-based chemical models enable sub-second verification of common operations. These simplified models are validated against high-fidelity references and physical experiments, with documented accuracy bounds informing safety margin requirements.

High-risk operations demand more sophisticated simulation. Physics-based models incorporating computational fluid dynamics, finite element analysis, and detailed reaction kinetics provide higher accuracy at greater computational cost. For operations involving novel materials, extreme conditions, or potential runaway scenarios, only high-fidelity simulation provides adequate basis for safety decisions. The computational overhead of detailed simulation is acceptable given the criticality of the operations being verified.

The connection between digital twin prediction and physical execution operates bidirectionally. Forward prediction evaluates proposed actions, while backward reconciliation compares predicted outcomes against observed results. Systematic divergence between prediction and observation indicates either model inadequacy or sensor degradation, triggering diagnostic procedures and potentially restricting operations until the discrepancy is resolved. This continuous validation maintains confidence in the digital twin's predictive accuracy throughout extended autonomous operation.

Advanced implementations incorporate uncertainty quantification into digital twin predictions, propagating parameter uncertainties through models to bound outcome distributions. Rather than predicting a single outcome, the system predicts a range of possible outcomes with associated probabilities, enabling risk-aware decision making that accounts for model limitations and measurement uncertainties~\cite{tayal2025a}.

\begin{figure}[t]
\centering
\includegraphics[width=0.88\textwidth]{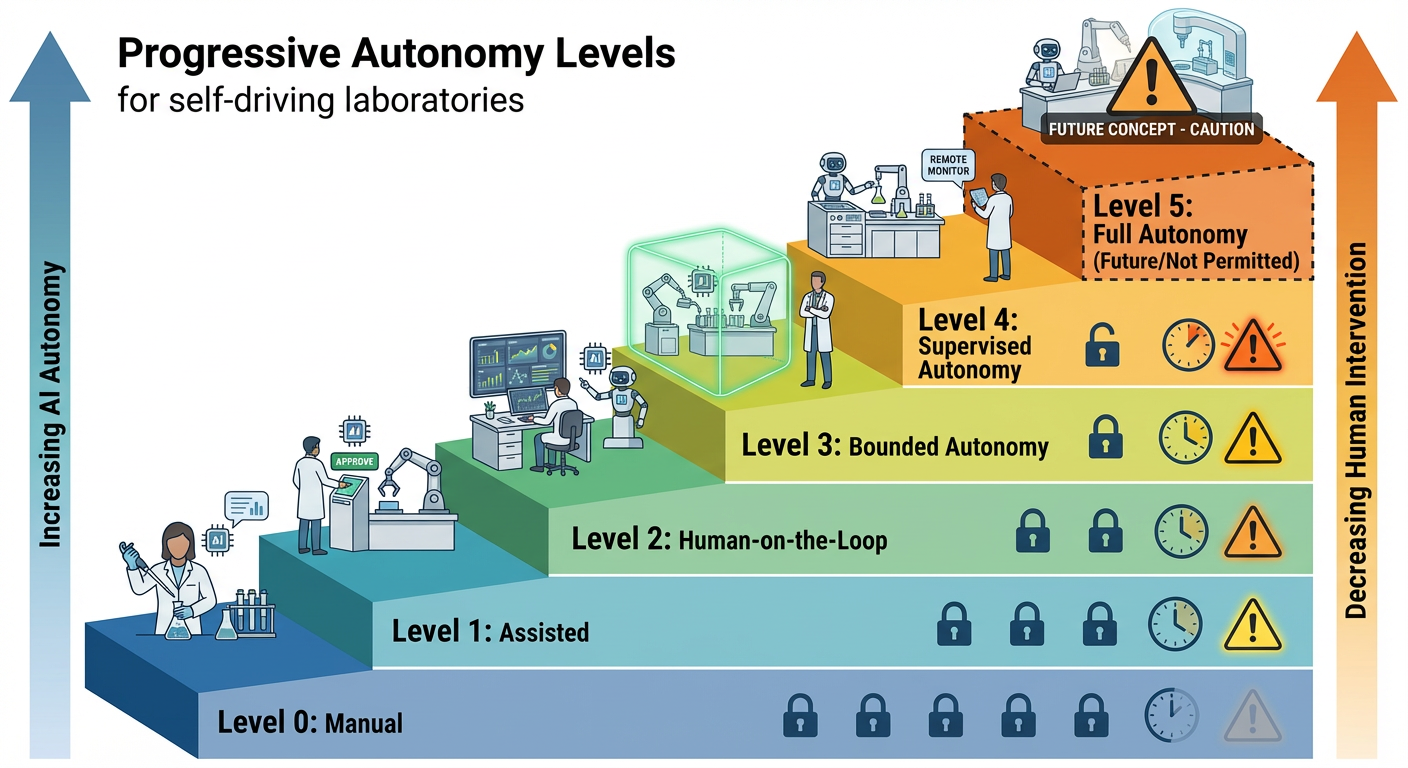}
\caption{\textbf{Progressive Autonomy Levels for Self-Driving Laboratories.} The framework defines six levels (0-5) representing increasing AI autonomy and decreasing human intervention. Level 0 (Manual) maintains full human control with AI as advisor; Level 1 (Assisted) requires human approval for each step; Level 2 (Human-on-the-Loop) enables AI execution under continuous human monitoring; Level 3 (Bounded Autonomy) permits autonomous operation within strictly defined ODDs with human standby; Level 4 (Supervised Autonomy) allows extended autonomous operation with periodic human oversight; Level 5 (Full Autonomy) represents the theoretical endpoint of full AI independence, depicted with a dotted outline to indicate this level is not currently achievable given the limitations of present AI systems and regulatory constraints. Left arrow shows increasing AI autonomy; right indicators show trust/verification requirements (lock icons), response time expectations (clocks), and risk tolerance (warning triangles). Color gradient transitions from cool blue (human-controlled) to warm orange (AI-autonomous), emphasizing the graduated transition of responsibility.}
\label{fig:progressive_autonomy}
\end{figure}

\subsection{Safety Instrumented Functions and Formal Verification}
\label{subsec:formal_verification}

Certain safety functions are sufficiently critical that probabilistic guarantees from testing and simulation provide inadequate assurance. For these functions, formal verification techniques provide mathematical proof of correctness, eliminating entire categories of potential failure modes~\cite{luckcuck2022formal}.

Formal verification approaches in SDL contexts typically focus on safety-critical software components: the Safety Kernel's decision logic, CBF implementations, and emergency shutdown sequences. Verification targets are selected based on criticality assessment, with effort concentrated on components whose failure could result in severe harm and which are amenable to formal analysis.

The Safe-ROS architecture~\cite{Benjumea_2025} demonstrates practical formal verification through its Safety Instrumented Function (SIF) framework. The architecture defines safety properties in temporal logic specifications, expressing requirements such as ``whenever an obstacle is detected within the safety radius, the robot must stop within the specified time bound.'' Model checking tools exhaustively verify that the control software satisfies these specifications for all possible input sequences within the modeled state space.

Formal verification provides strongest guarantees for components with limited complexity and well-defined interfaces. The strategy of concentrating critical safety logic in small, formally verified components while allowing more flexibility in non-critical components represents a practical accommodation to verification scalability limitations. The verified Safety Kernel provides a trusted foundation upon which more complex but less critical functionality can be constructed.

The verification process itself produces valuable artifacts beyond correctness proofs. Counter-examples generated during failed verification attempts reveal edge cases and boundary conditions that might escape conventional testing. Invariant properties identified during verification often expose implicit assumptions that merit documentation and monitoring. The discipline of specifying precise formal requirements frequently identifies ambiguities or conflicts in informal safety requirements before implementation begins.

\subsection{Human-in-the-Loop Governance and Progressive Autonomy}
\label{subsec:progressive_autonomy}

Technical safety mechanisms operate within governance frameworks that define the appropriate scope and conditions for autonomous operation. Effective governance balances the scientific benefits of autonomy against risk, progressively expanding autonomous capability as systems demonstrate safe operation within established boundaries.

The progressive autonomy model, illustrated in Figure~\ref{fig:progressive_autonomy}, defines levels of human involvement ranging from full manual control through fully autonomous operation. At lower autonomy levels, human operators maintain direct control over all safety-critical decisions, with AI systems serving in advisory roles. Intermediate levels permit autonomous operation within constrained domains, with automatic escalation to human oversight when operations approach domain boundaries or encounter novel situations. Higher autonomy levels reduce human involvement to monitoring and exception handling, with the system managing routine operations independently.

Transitions between autonomy levels should follow demonstrated competence criteria rather than arbitrary timelines. A system operating at Level 2 (human-on-the-loop) advances to Level 3 (bounded autonomy) only after accumulating sufficient incident-free operational history (typically hundreds of successful operations) to statistically validate safety performance claims. Regression to lower autonomy levels occurs automatically upon safety incidents or constraint violations, with advancement resuming only after root cause analysis and corrective action verification.

The Osprey framework~\cite{hellert2025ospreyproductionreadyagenticai} implements progressive autonomy through its ``Plan-First'' architecture, requiring explicit human approval for plans before hardware interaction. Generated execution plans include complete dependency graphs, resource requirements, and explicit safety assessments with confidence metrics, enabling informed human review of proposed operations. This transparency supports meaningful human oversight rather than perfunctory approval of opaque AI decisions.

Communication design significantly impacts governance effectiveness. Safety-relevant information must be presented in forms that support human comprehension and decision-making within available time constraints. Information overload---presenting operators with excessive data streams requiring continuous monitoring---produces vigilance decrements that undermine the oversight function. Effective interfaces highlight anomalies and decision points while suppressing routine status information, maintaining operator engagement for genuinely novel situations.

The progressive autonomy model provides a framework for responsible deployment that balances innovation with safety. Organizations can begin at lower autonomy levels with extensive human oversight, gradually expanding autonomous capability as systems demonstrate reliable safe operation. This measured approach builds both technical confidence through operational validation and organizational trust through demonstrated performance.

\subsection{Laboratory Equipment Operation Protocols}
\label{subsec:equipment_protocols}

Autonomous laboratory systems must interface with diverse laboratory equipment through standardized protocols that ensure both operational correctness and safety compliance. This section documents the procedural sequences for three critical classes of laboratory equipment: pipetting systems, liquid dispensing apparatus, and robotic arm coordination with mixing devices. These procedures exemplify the detailed operational protocols that must be formally specified before AI-driven autonomous operation, illustrating how procedural complexity at the execution level requires explicit representation in planning and control systems..

\subsubsection{Pipetting Equipment Operation Sequence}

\begin{figure*}[!htbp]
\centering
\setlength{\tabcolsep}{1pt}
\renewcommand{\arraystretch}{0.6}
\newlength{\imgheight}
\setlength{\imgheight}{0.13\textheight}
\begin{tabular}{@{}c@{\hspace{3pt}}c@{\hspace{3pt}}c@{}}
% Row 1: Three pipetting images with fixed height (no keepaspectratio for uniform size)
\includegraphics[width=0.32\textwidth,height=\imgheight]{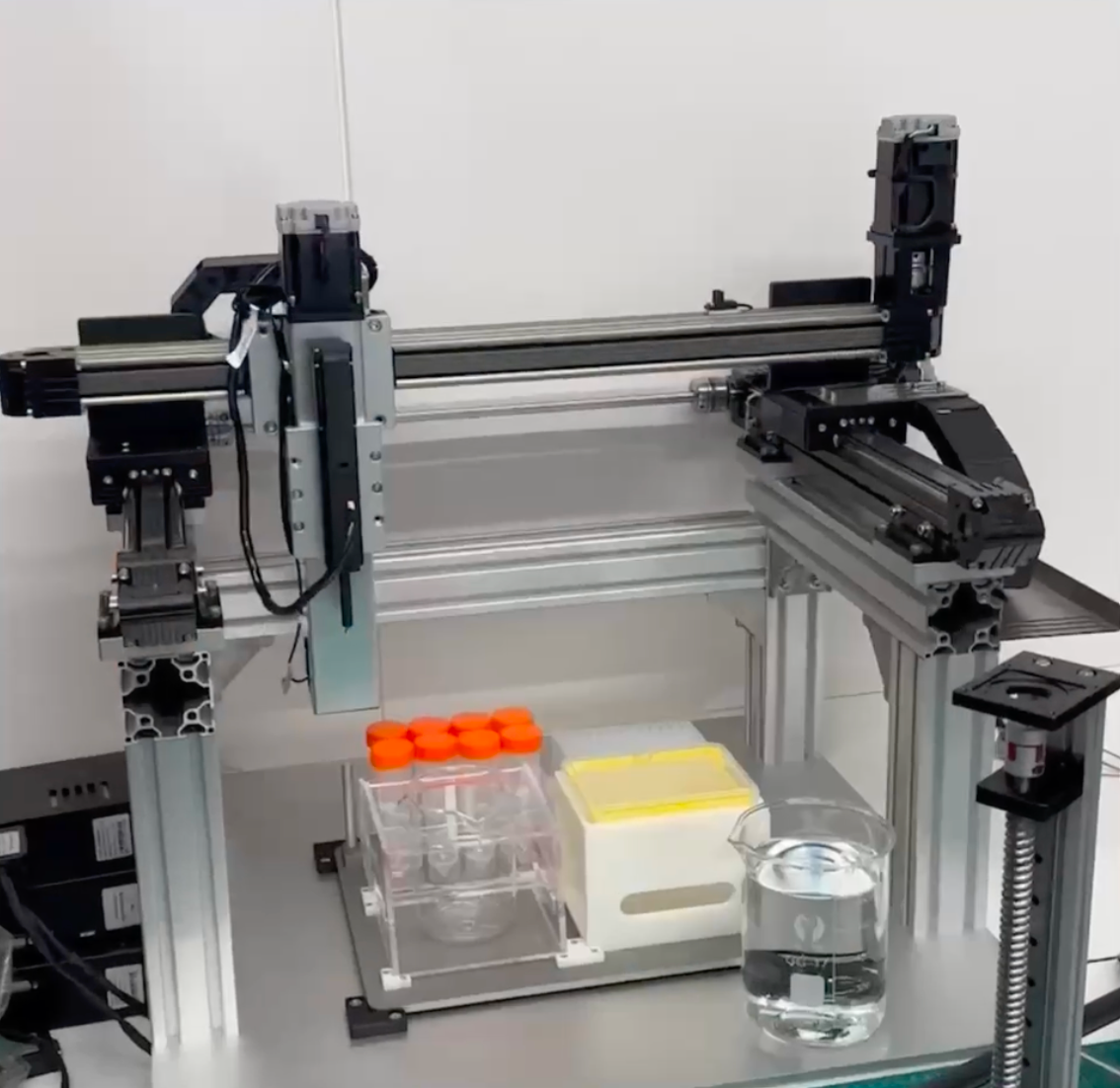} &
\includegraphics[width=0.32\textwidth,height=\imgheight]{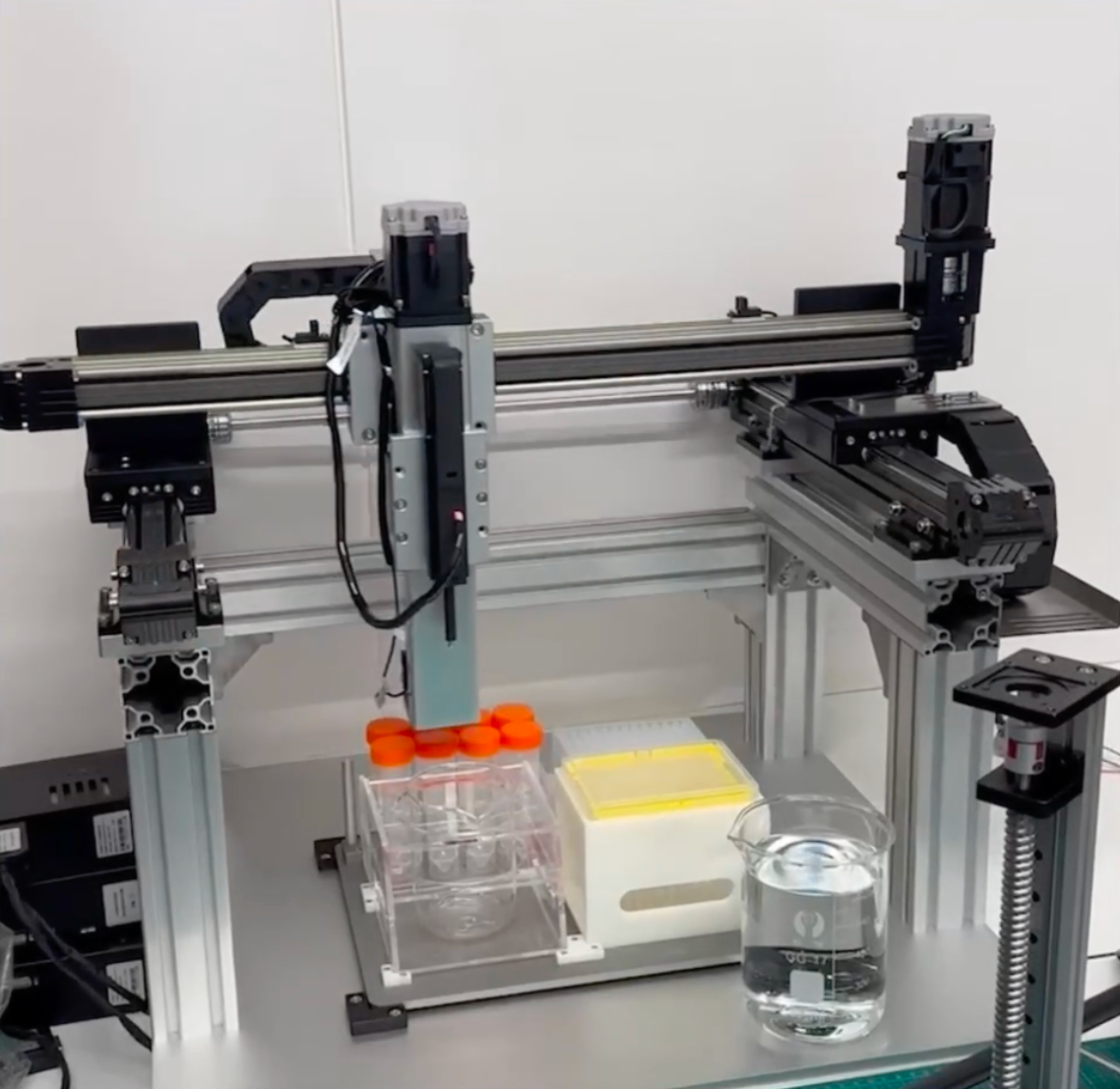} &
\includegraphics[width=0.32\textwidth,height=\imgheight]{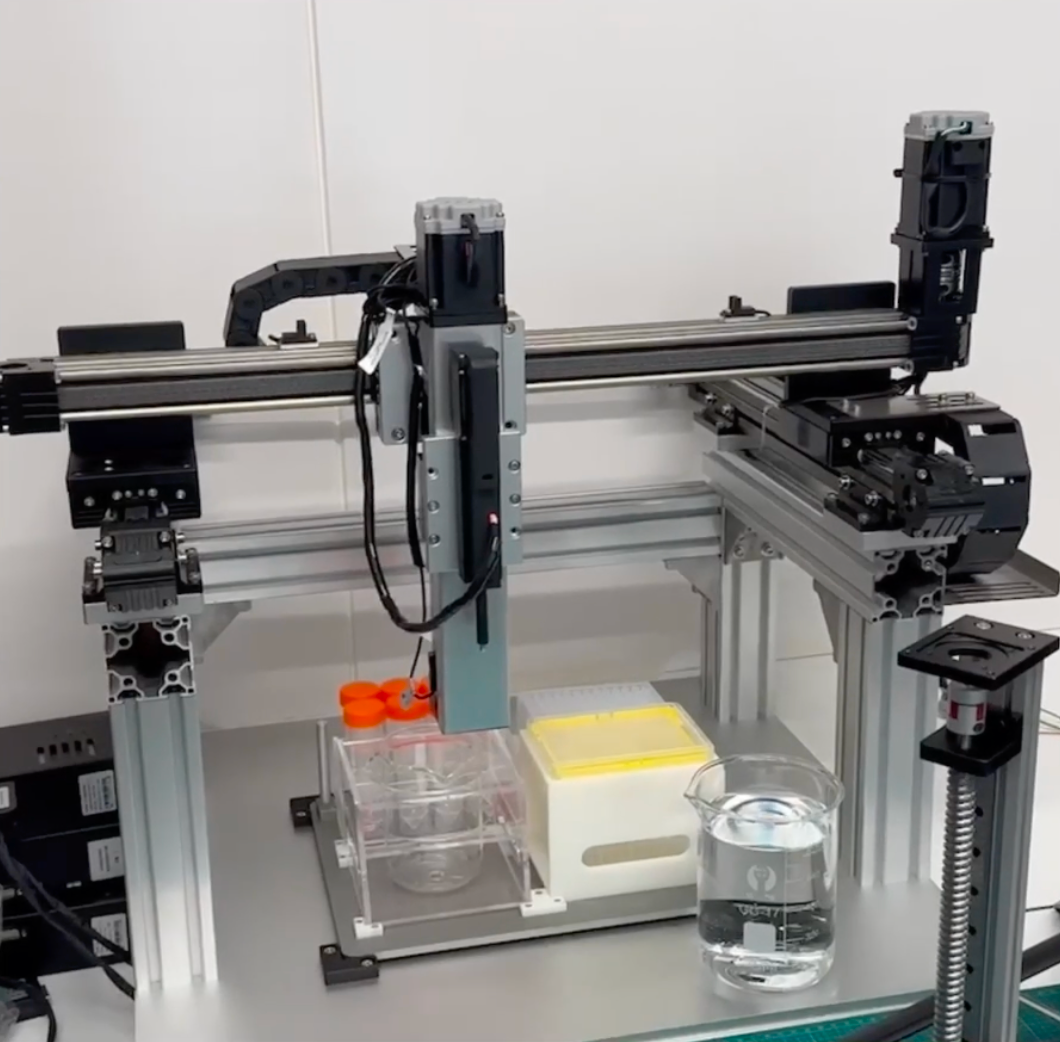} \\[1pt]
\small (a) Move x-axis forward & \small (b) Move y-axis forward & \small (c) Move z-axis forward \\[2pt]
% Row 2: Three more pipetting images with fixed height
\includegraphics[width=0.32\textwidth,height=\imgheight]{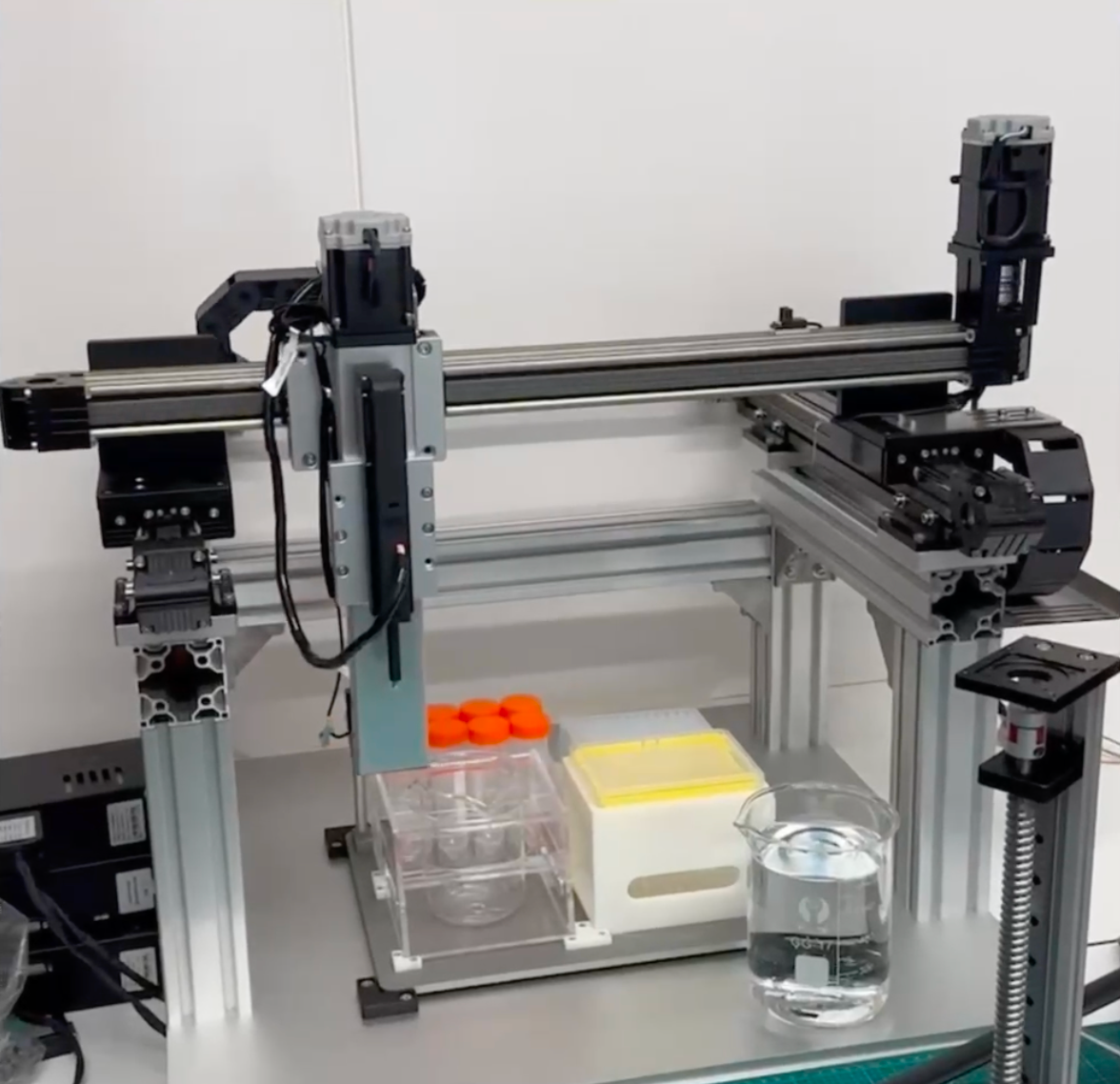} &
\includegraphics[width=0.32\textwidth,height=\imgheight]{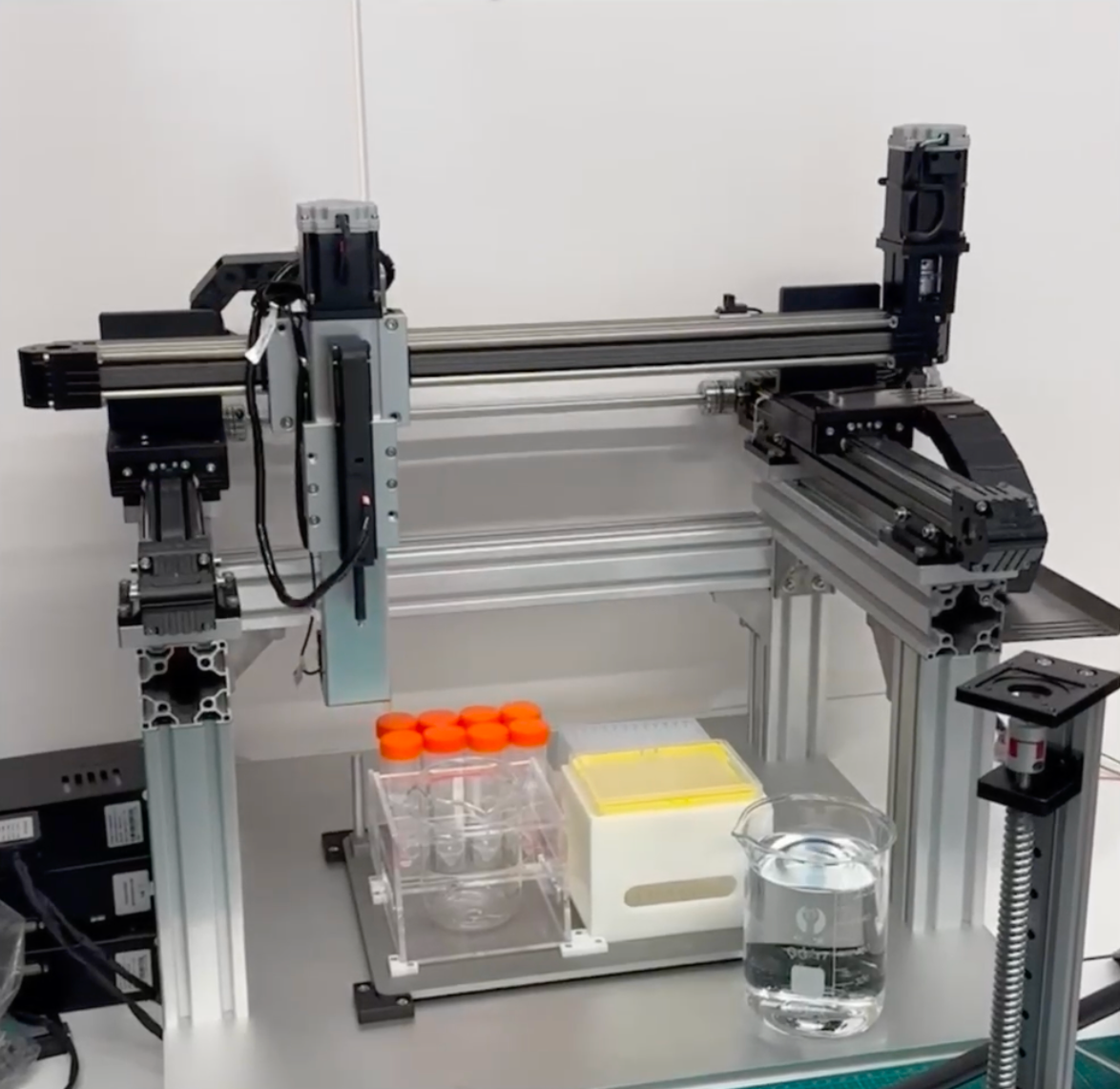} &
\includegraphics[width=0.32\textwidth,height=\imgheight]{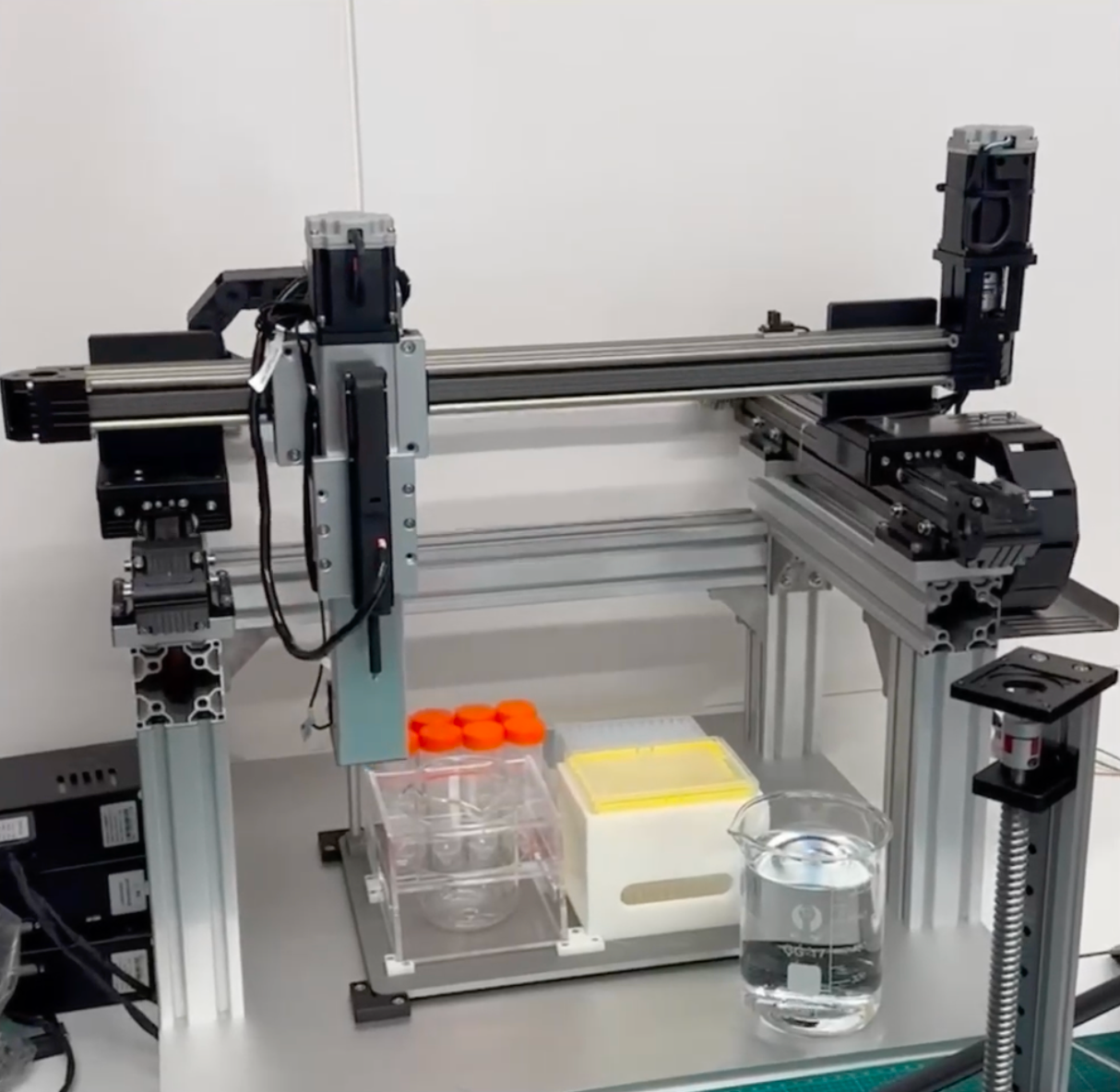} \\[1pt]
\small (d) Move backward along the z-axis & \small (e) Move backward along the y-axis & \small (f) Move backward along the x-axis \\
\end{tabular}
\vspace{3pt}
\caption{\textbf{Complete Pipetting Equipment Operation Sequence.}
\textbf{(a)} Move x-axis forward: horizontal translation toward the target vessel or well array.
\textbf{(b)} Move y-axis forward: lateral positioning to align the pipette tip with the intended source or destination.
\textbf{(c)} Move z-axis forward: vertical descent for tip immersion (aspiration) or approach to dispense surface.
\textbf{(d)} Move backward along the z-axis: vertical retraction after aspiration or after dispense, clearing the liquid surface.
\textbf{(e)} Move backward along the y-axis: lateral withdrawal from the vessel or well.
\textbf{(f)} Move backward along the x-axis: return translation to home position or to the next source/destination. The sequence (a)$\rightarrow$(f) defines the canonical motion pattern for a single transfer; errors in any axis motion propagate to volumetric inaccuracy or collision risk.}
\label{fig:pipetting_complete}
\end{figure*}

Pipetting operations represent fundamental activities in biological and chemical laboratories, involving precise volumetric transfer of microliter quantities of liquids. The automated execution of pipetting operations requires that the AI planning system understand not only the desired volume transfer but the correct sequence of mechanical motions along each axis that achieve accurate, reproducible, and collision-free transfers. Figure~\ref{fig:pipetting_complete} illustrates the complete procedural sequence as a canonical axis-motion pattern: approach along x, y, and z (a--c), then retract along z, y, and x (d--f).

The sequence (a)$\rightarrow$(f) makes explicit the procedural complexity that AI planning systems must handle. The approach phase (a--c)---moving the pipette forward along x, then y, then z---positions the tip at the source or destination with correct immersion depth and alignment; the retraction phase (d--f)---withdrawing along z, then y, then x---avoids tip collision and cross-contamination. While conceptually simple (``transfer 10 µL from vessel A to vessel B''), each axis motion involves explicit parameters: velocity profiles, acceleration limits, and positional bounds. These specifications cannot be inferred from natural language; they must be explicitly taught to AI systems or enforced through hardware-level constraints.

This procedural formalization directly addresses a key element of the Syntax-to-Safety Gap introduced in Section~\ref{sec:introduction}: a syntactically correct AI command (``perform pipetting operation'') must map to a correctly ordered axis-motion sequence (e.g., x-, y-, z-forward then z-, y-, x-backward) with documented timing and safety margins. The gap between these representations constitutes a critical failure point where AI understanding and laboratory protocol knowledge must intersect precisely.

\subsubsection{Liquid Dispenser Operation and Precision Delivery}

\begin{figure*}[!htbp]
\centering
\setlength{\tabcolsep}{1pt}
\renewcommand{\arraystretch}{0.6}
\setlength{\imgheight}{0.13\textheight}
\begin{tabular}{@{}c@{\hspace{3pt}}c@{\hspace{3pt}}c@{}}
% Row 1: Three dispenser images with fixed height for compact layout
\includegraphics[width=0.32\textwidth,height=\imgheight]{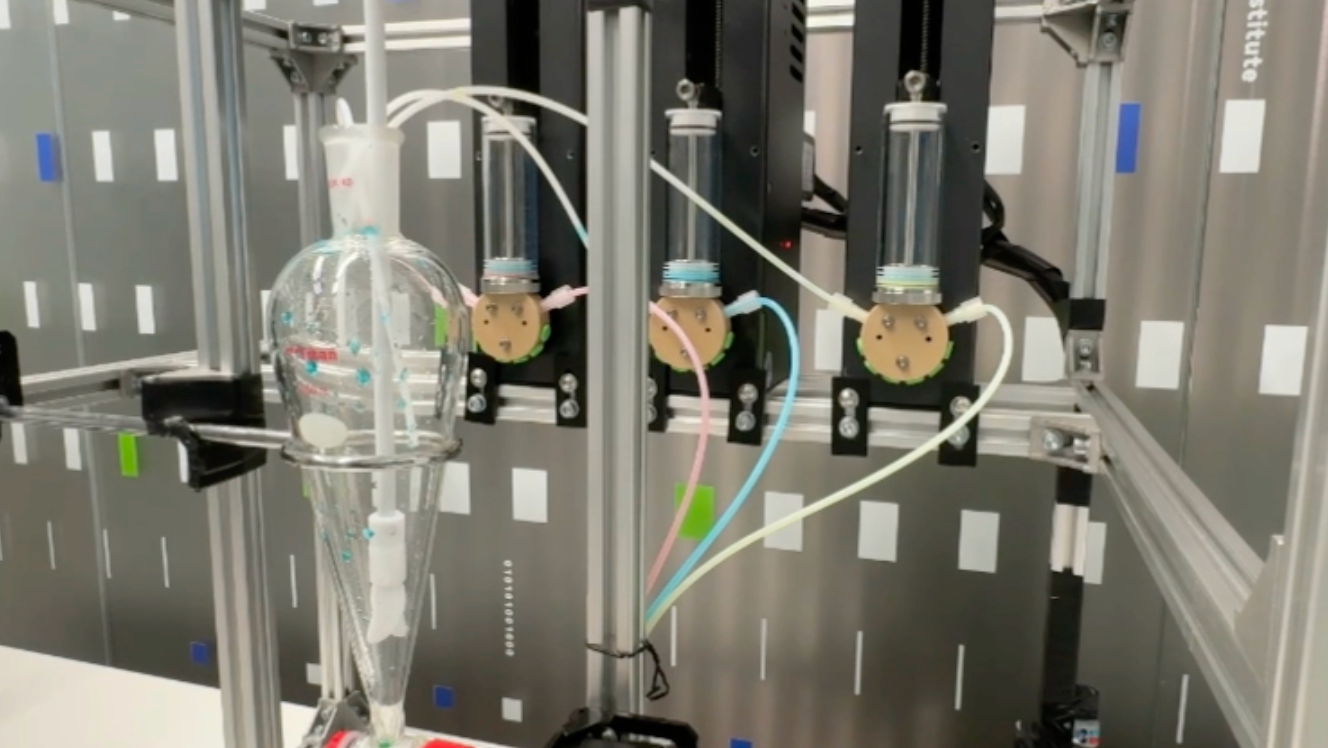} &
\includegraphics[width=0.32\textwidth,height=\imgheight]{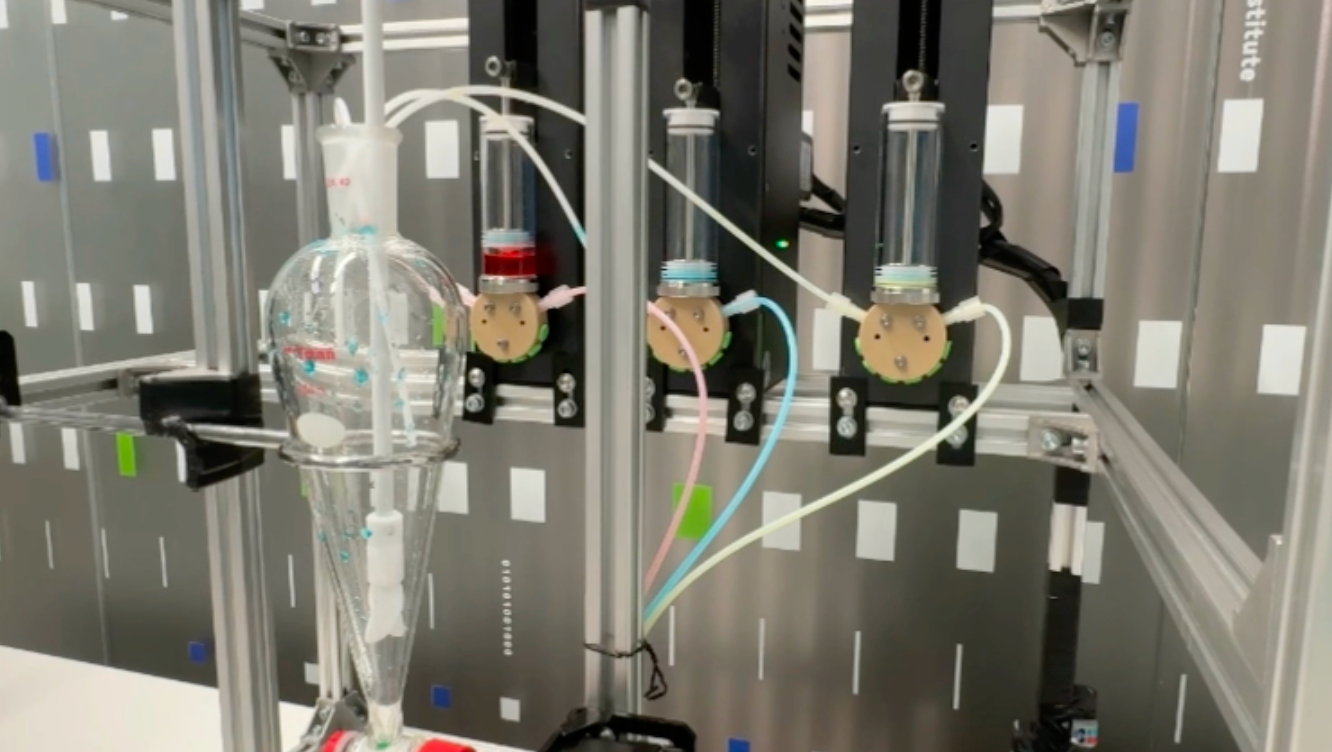} &
\includegraphics[width=0.32\textwidth,height=\imgheight]{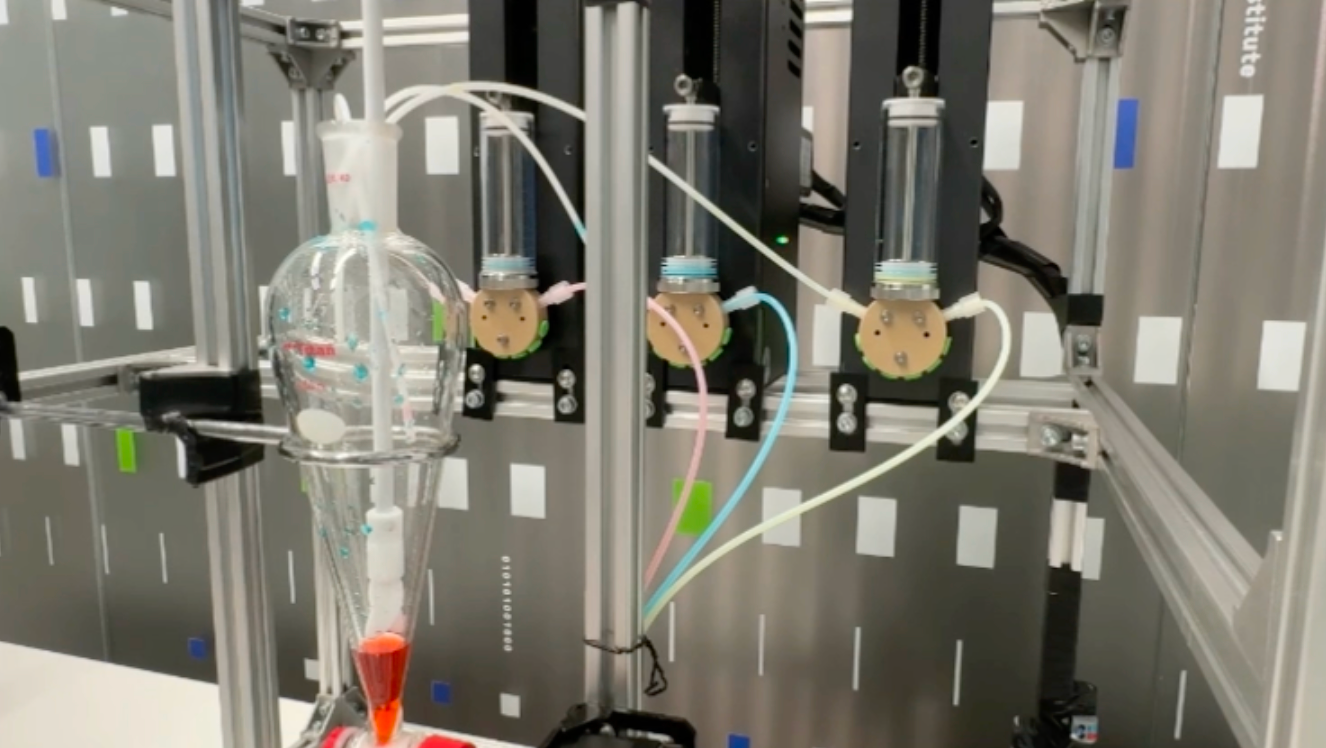} \\[1pt]
\small (i) Connect reservoir to dispenser & \small (ii) Extract red liquid & \small (iii) Inject into the separating funnel \\[2pt]
% Row 2: Three more dispenser images
\includegraphics[width=0.32\textwidth,height=\imgheight]{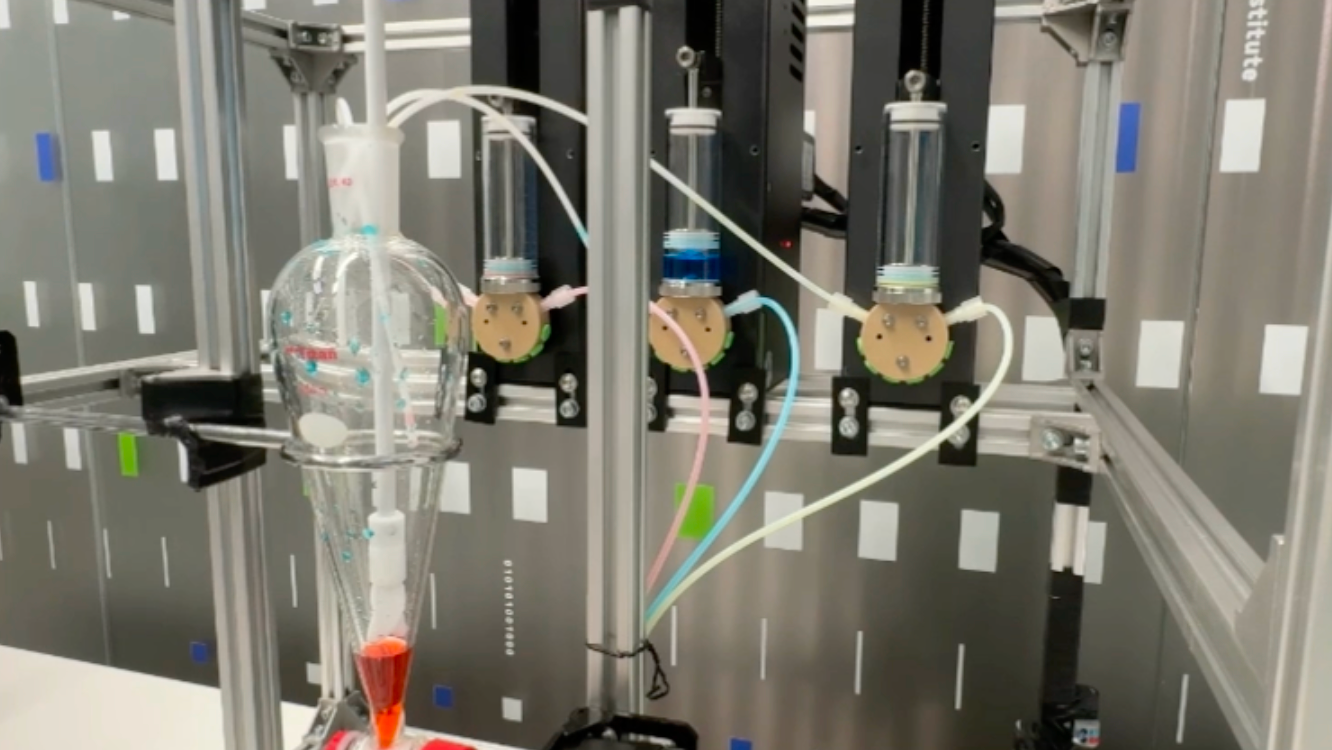} &
\includegraphics[width=0.32\textwidth,height=\imgheight]{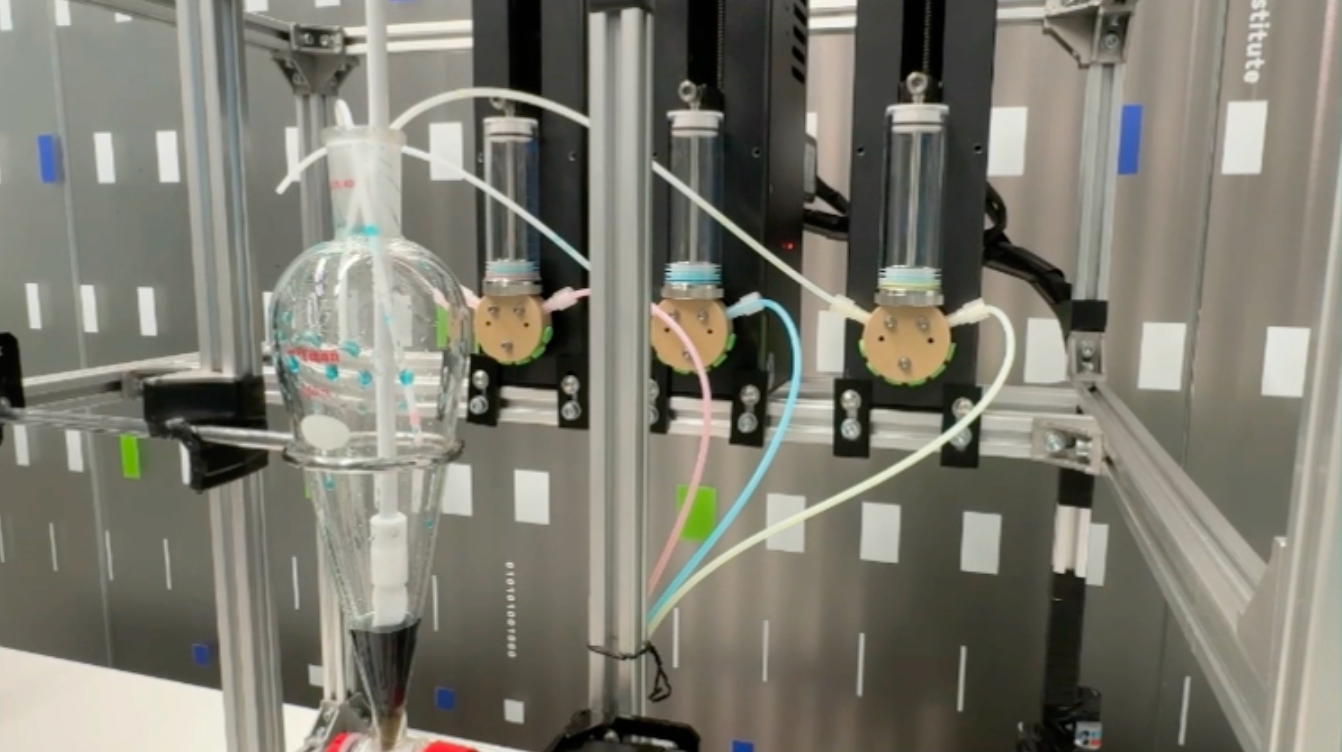} &
\includegraphics[width=0.32\textwidth,height=\imgheight]{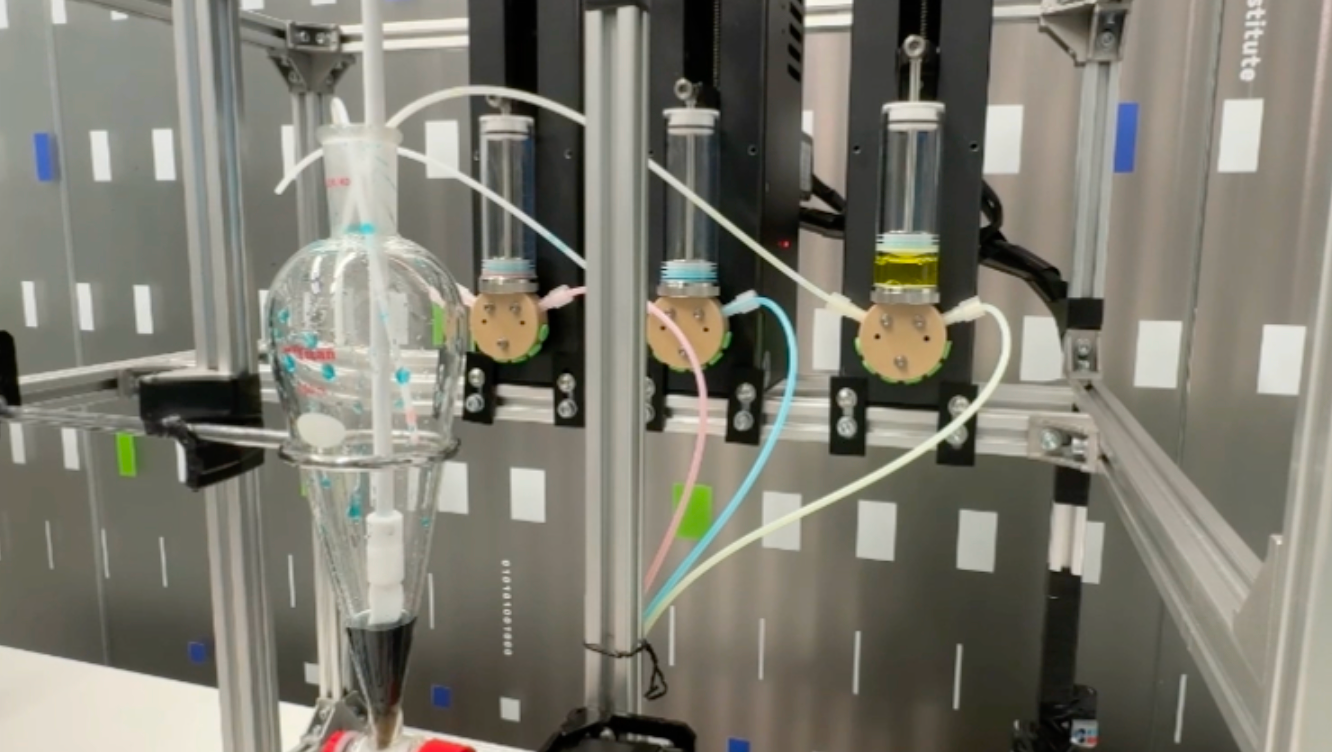} \\[1pt]
\small (iv) Extract blue liquid & \small (v) Inject into the separating funnel & \small (vi) Extract green liquid \\
\end{tabular}
\vspace{3pt}
\caption{\textbf{Complete Liquid Dispensing System Operation Sequence.}
\textbf{(i)} Connect reservoir to dispenser: secure tubing or syringe connection, prime lines to expel air, and verify pressure/flow calibration.
\textbf{(ii)} Extract red liquid: draw the first reagent from the reservoir into the dispenser under controlled flow and volume monitoring.
\textbf{(iii)} Inject into the separating funnel: deliver the red liquid into the central vessel (separating funnel) with completion detection (e.g., optical or gravimetric).
\textbf{(iv)} Extract blue liquid: switch or connect to the second reservoir and extract the next reagent.
\textbf{(v)} Inject into the separating funnel: deliver the blue liquid into the same separating funnel for layered or sequential processing.
\textbf{(vi)} Extract green liquid: connect to the third reservoir and extract the final reagent; the sequence (i)$\rightarrow$(vi) exemplifies multi-reagent coordination and contamination-aware ordering.}
\label{fig:dispenser_complete}
\end{figure*}

Beyond standard pipetting, liquid dispensing systems (peristaltic pumps, syringe dispensers, and pressure-driven systems) enable precise delivery of larger volumes and specialized solutions while maintaining sterility and reproducibility. Figure~\ref{fig:dispenser_complete} illustrates a complete operational sequence: connecting the reservoir to the dispenser (i), then alternating extraction and injection for multiple reagents---red liquid extracted (ii) and injected into the separating funnel (iii), blue liquid extracted (iv) and injected (v), and green liquid extracted (vi)---exemplifying multi-reagent coordination into a single vessel.

\subsubsection{Robotic Arm and Stirrer Coordinated Operation}

\begin{figure*}[!htbp]
\centering
\setlength{\tabcolsep}{1pt}
\renewcommand{\arraystretch}{0.6}
\newlength{\armfigheight}
\setlength{\armfigheight}{0.18\textheight}
\begin{tabular}{@{}c@{\hspace{3pt}}c@{}}
% Row 1: two images
\includegraphics[width=0.48\textwidth,height=\armfigheight]{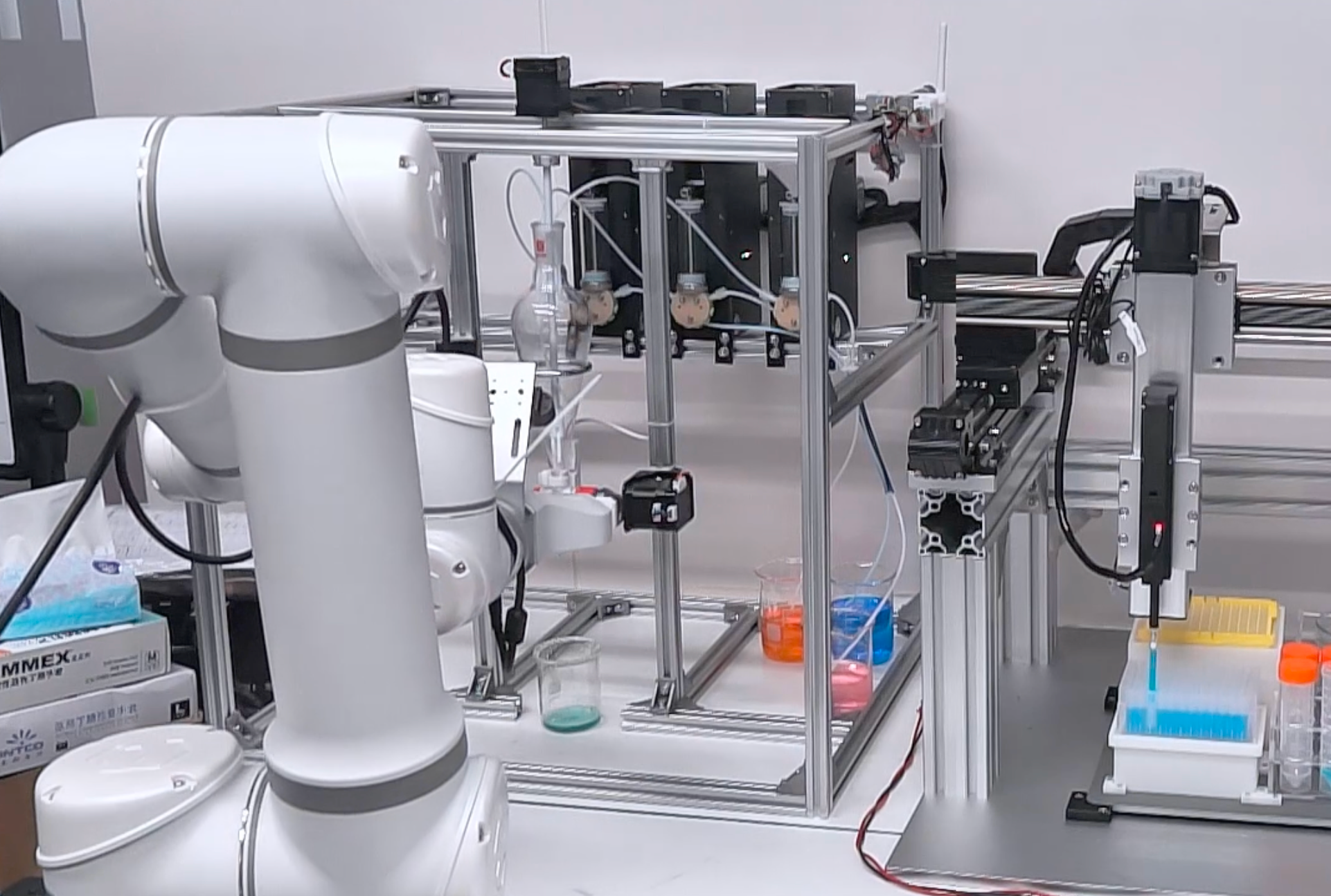} &
\includegraphics[width=0.48\textwidth,height=\armfigheight]{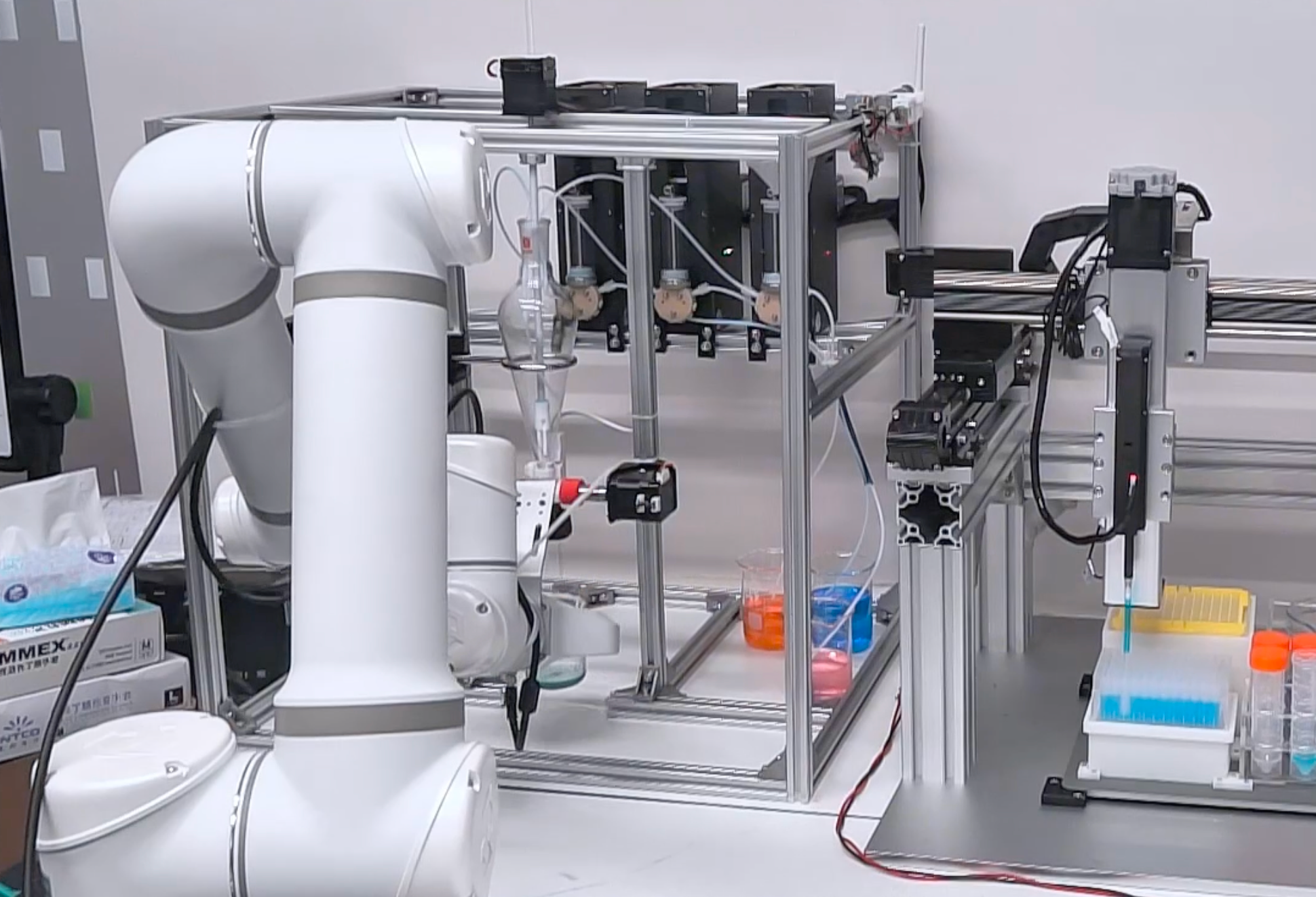} \\[1pt]
\small (a) Arm approaching vessel & \small (b) Reagent sampling \\[4pt]
% Row 2: two more images
\includegraphics[width=0.48\textwidth,height=\armfigheight]{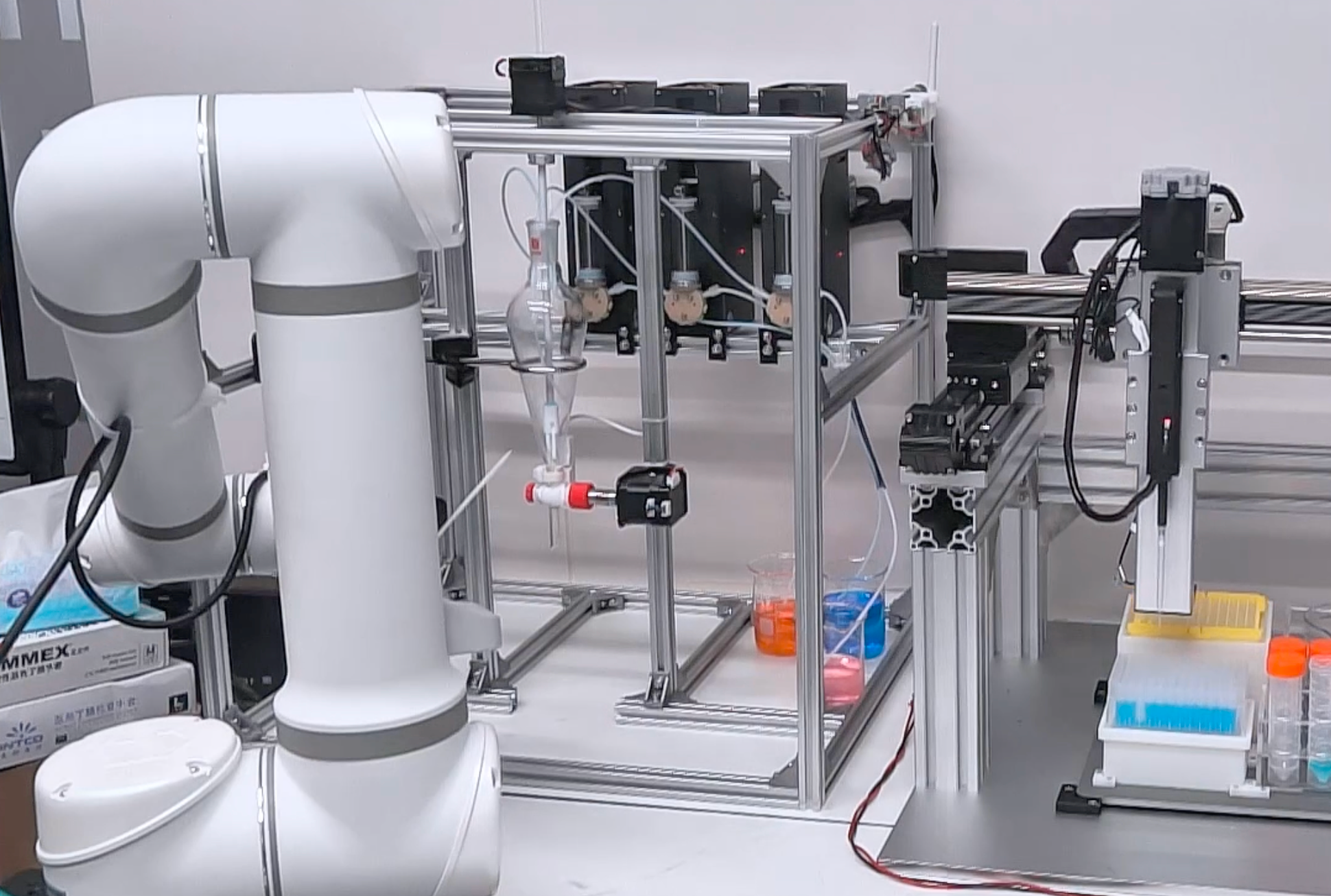} &
\includegraphics[width=0.48\textwidth,height=\armfigheight]{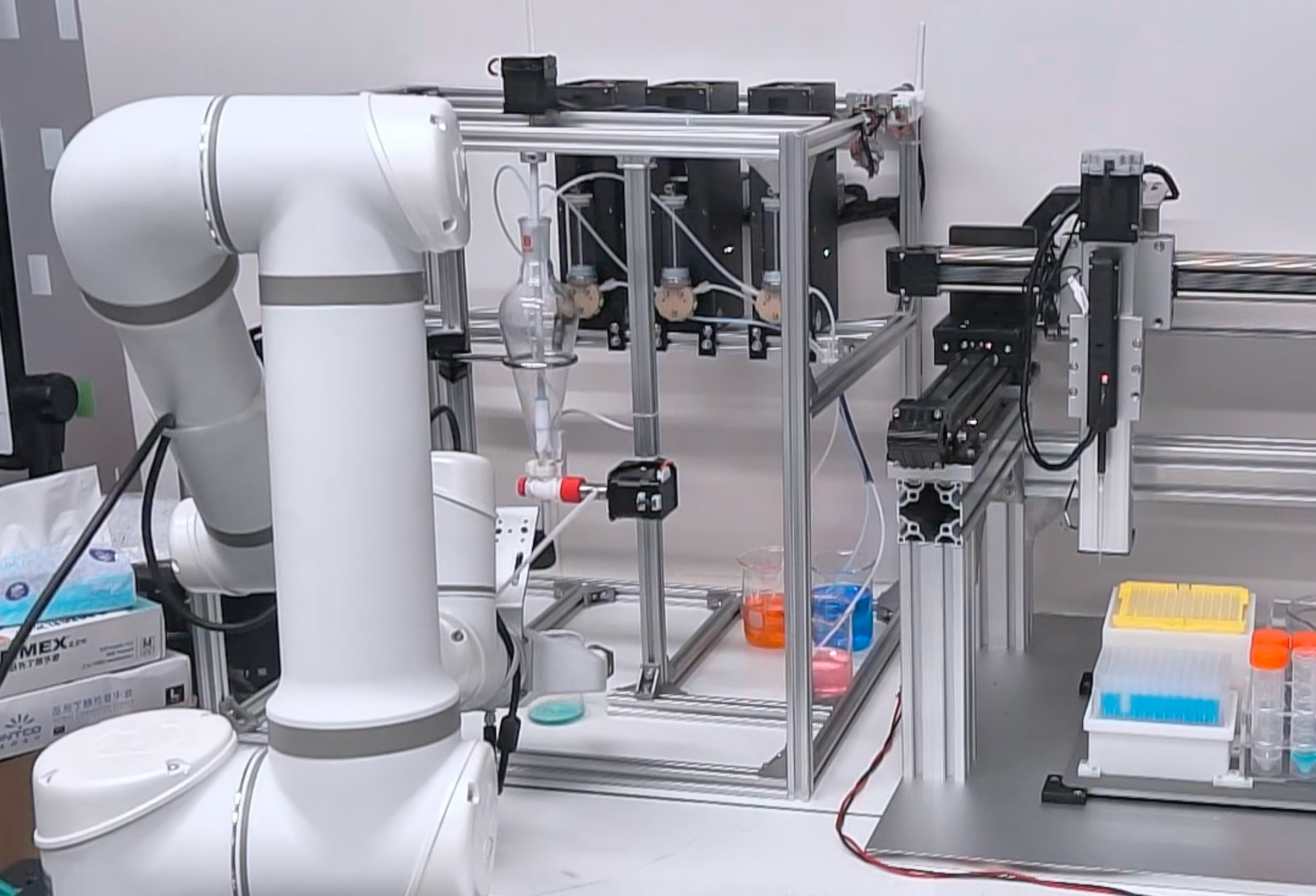} \\[1pt]
\small (c) Arm repositioning & \small (d) Completion or next step \\
\end{tabular}
\vspace{8pt}
\caption{\textbf{Robotic Arm and Stirrer Coordinated Operation.}
\textbf{(a)} Arm approaching vessel: initial positioning with stirrer active in the vessel.
\textbf{(b)} Reagent sampling: arm performs pipetting or sampling while stirrer maintains mixing.
\textbf{(c)} Arm repositioning: trajectory avoids interference with stirrer and vessel.
\textbf{(d)} Completion or next step: arm retracts or moves to next target.}
\label{fig:arm_stirrer}
\end{figure*}

When a robotic arm and a stirrer share the same workspace, coordination is essential so that both can perform their tasks without collision or interference. Figure~\ref{fig:arm_stirrer} illustrates a sequence in which the robotic arm and stirrer operate concurrently: the arm performs reagent handling or sampling while the stirrer maintains mixing in the vessel. Such coordination relies on spatial and temporal planning that respects safety boundaries (e.g., CBF-based collision avoidance as in Section~\ref{subsec:cbf}).

% \begin{figure*}[!htbp]
% \centering
% \setlength{\tabcolsep}{2pt}
% \newlength{\armfigheight}
% \setlength{\armfigheight}{3.8cm}
% \begin{tabular}{@{}c@{\hspace{2pt}}c@{}}
% \begin{minipage}[c]{0.48\textwidth}
%   \centering
%   \includegraphics[height=\armfigheight, keepaspectratio]{assets/f22.png}\\[2pt]
%   \small (a) Arm approaching vessel
% \end{minipage} &
% \begin{minipage}[c]{0.48\textwidth}
%   \centering
%   \includegraphics[height=\armfigheight, keepaspectratio]{assets/f23.png}\\[2pt]
%   \small (b) Reagent sampling
% \end{minipage} \\[6pt]
% \begin{minipage}[c]{0.48\textwidth}
%   \centering
%   \includegraphics[height=\armfigheight, keepaspectratio]{assets/f24.png}\\[2pt]
%   \small (c) Arm repositioning
% \end{minipage} &
% \begin{minipage}[c]{0.48\textwidth}
%   \centering
%   \includegraphics[height=\armfigheight, keepaspectratio]{assets/f25.png}\\[2pt]
%   \small (d) Completion or next step
% \end{minipage}
% \end{tabular}
% \vspace{4pt}
% \caption{\textbf{Robotic Arm and Stirrer Coordinated Operation.}
% \textbf{(a)} Arm approaching vessel: initial positioning with stirrer active in the vessel.
% \textbf{(b)} Reagent sampling: arm performs pipetting or sampling while stirrer maintains mixing.
% \textbf{(c)} Arm repositioning: trajectory avoids interference with stirrer and vessel.
% \textbf{(d)} Completion or next step: arm retracts or moves to next target. Four frames exemplify collision-free multi-device operation in SDL environments.}
% \label{fig:arm_stirrer}
% \end{figure*}

The sequences in Figures~\ref{fig:pipetting_complete}, \ref{fig:dispenser_complete}, and \ref{fig:arm_stirrer} reveal the procedural richness underlying laboratory automation. For the dispenser, the ordering (i)$\rightarrow$(vi) encodes reservoir connection, then repeated extract-inject cycles (red, blue, green) into the separating funnel; each step requires priming, flow monitoring, and completion detection to avoid air bubbles, overfill, or cross-contamination. While higher-level AI planning might conceptualize the task as ``add reagents A, B, and C to the funnel,'' the actual realization requires the correct mechanical sequence and contamination-aware ordering.

The key insight for Safe-SDL architecture is that this procedural complexity cannot be entirely hidden from higher-level planning systems. The AI planner must understand not only that an operation is desired but the prerequisites (e.g., reservoir connected, lines primed), the correct step order (connect $\rightarrow$ extract $\rightarrow$ inject, per reagent), the error modes (e.g., air in line, overflow), and the recovery strategies for each failure. This knowledge enables realistic plan generation that accounts for equipment capabilities and limitations, identifies infeasible operation sequences before commencing physical execution, and specifies recovery behaviors for plausible failure scenarios.

\section{Framework Validation and Application Scenarios}
\label{sec:evaluation}

Validating the Safe-SDL framework requires demonstrating that its theoretical principles---Operational Design Domains, Control Barrier Functions, and transactional protocols---can be effectively instantiated in real autonomous laboratory systems and that these instantiations provide measurable safety improvements. This section validates the framework through multiple complementary approaches: assessing how existing benchmarks reveal the necessity of architectural safety mechanisms (Section~\ref{subsec:benchmarks}), examining formal verification techniques applicable to Safety Kernel components (Section~\ref{subsec:verification_limits}), analyzing operational evidence from SDL deployments where Safe-SDL principles have been applied (Section~\ref{subsec:incident_analysis}), and demonstrating domain-specific implementations of the framework in chemistry, biology, and materials science contexts (Section~\ref{subsec:case_studies}). Collectively, these validation approaches establish both the theoretical soundness and practical viability of Safe-SDL.

\subsection{Benchmarking the Necessity of Architectural Safety}
\label{subsec:benchmarks}

To validate the necessity of Safe-SDL's architectural approach, we examine foundation model performance on laboratory safety benchmarks. If models reliably made safe decisions, architectural safety mechanisms would be unnecessary; benchmark results demonstrating systematic safety failures justify the defensive architecture we propose. The LabSafety Bench~\cite{li2024labsafety} and SOSBENCH~\cite{zhang2025sosbench} provide standardized evaluation protocols that reveal the gap between model capabilities and safety requirements.

The LabSafety Bench~\cite{li2024labsafety} establishes an important 
foundation for safety evaluation in scientific contexts. 
This benchmark presents language models with laboratory scenarios requiring 
safety-relevant decisions, evaluating the models' ability to identify hazards, 
select appropriate responses, and avoid dangerous recommendations. 
The benchmark encompasses multiple question formats including hazard identification, 
procedure evaluation, and emergency response selection, enabling assessment of 
different safety-relevant capabilities.

\begin{figure}[t]
\centering
\includegraphics[width=0.95\textwidth]{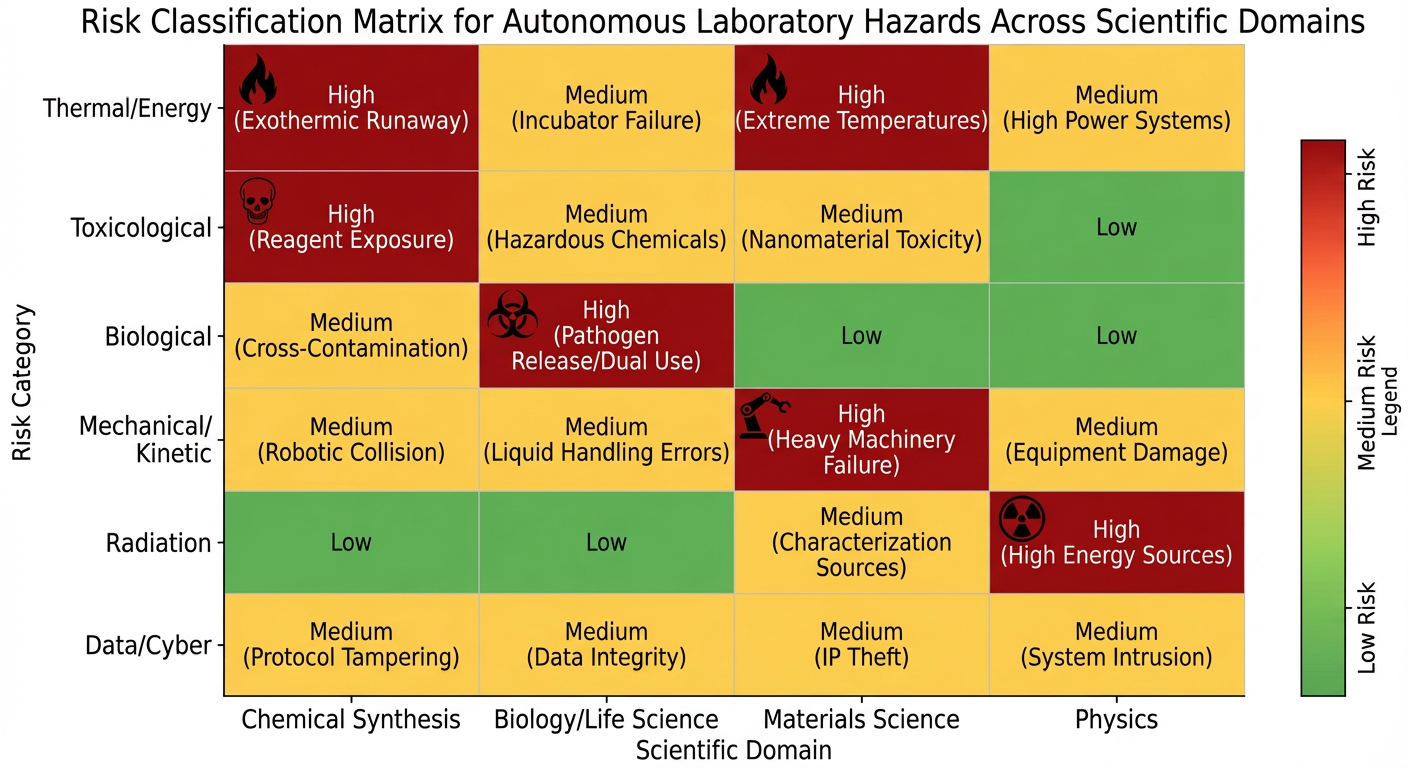}
\caption{\textbf{Domain-Specific Risk Classification Matrix.} 
This heatmap illustrates risk levels across six hazard categories (rows) and 
four scientific domains (columns), adapted from the safety taxonomy and 
evaluation framework established in \cite{li2024labsafety}. 
Cell colors indicate risk severity using a gradient from green (low risk) 
through yellow (medium) to red (high risk). 
High-risk cells (dark red) include Chemical-Thermal, Chemical-Toxicological, 
Biology-Biological, Materials-Thermal, and Materials-Mechanical hazards. 
Representative icons mark key cells: flame symbols for thermal hazards, 
skull for toxicological risks, biohazard symbols for biological dangers, 
and robot arms for mechanical/kinetic hazards. 
The matrix demonstrates that each scientific domain presents a characteristic 
hazard profile requiring tailored safety mechanisms. 
The color scale legend on the right enables quantitative interpretation of risk levels. 
This risk landscape analysis informs the domain-specific adaptations 
discussed in Section~\ref{subsec:case_studies}.}
\label{fig:risk_matrix}
\end{figure}

Figure~\ref{fig:risk_matrix} provides a systematic visualization of the risk landscape 
across scientific domains and hazard categories, based on the cross-domain 
safety taxonomy developed by Zhou et al.~\cite{li2024labsafety}. 
Benchmark results demonstrate substantial performance gaps, with foundation models frequently failing to identify hazardous situations during unsupervised operation in complex laboratory scenarios~\cite{li2024labsafety}. 
These findings validate Safe-SDL's core premise: architectural safety mechanisms are essential because current AI capabilities remain inadequate for unsupervised safety-critical decision making in laboratory contexts. The Syntax-to-Safety Gap is not a hypothetical concern but a documented empirical reality.

System-level validation of Safe-SDL requires demonstrating that architectural safety mechanisms---ODD validation, CBF enforcement, CRUTD protocols---successfully intercept unsafe commands that foundation models generate. The case studies in Section~\ref{subsec:case_studies} provide such demonstrations, showing how Safe-SDL principles implemented in systems like UniLabOS prevent hazardous operations that would otherwise proceed to execution. The SOSBENCH~\cite{zhang2025sosbench} benchmark complements LabSafety Bench by assessing domain-specific technical knowledge, further establishing that safety cannot rely on model capabilities alone.

\subsection{Formal Verification of Safety-Critical Components}
\label{subsec:verification_limits}

Safe-SDL's Safety Kernel and CBF components are amenable to formal verification, providing mathematical guarantees that complement empirical validation. We examine how verification techniques apply to our framework's critical components and acknowledge the inherent scope limitations. Verification validates that correctly implemented CBFs provably maintain system states within safe sets (Section~\ref{subsec:cbf}), that ODD constraint checking exhaustively covers all boundary conditions, and that CRUTD state transitions preserve safety invariants.

The verified components in architectures such as Safe-ROS~\cite{Benjumea_2025} 
typically constitute a small fraction of total system code. 
While critical safety logic may be formally verified, supporting infrastructure 
including sensor drivers, communication protocols, and orchestration logic 
often relies on conventional testing. 
The composition of verified and unverified components requires careful 
interface specification to preserve safety properties across component boundaries.

Model checking approaches face state space explosion challenges that limit 
the complexity of verifiable systems. 
Verification of continuous-time control systems requires abstraction to discrete 
approximations, introducing potential gaps between verified models and 
implemented behavior. 
Hybrid system verification techniques address some of these limitations but 
remain computationally intensive for systems of practical scale.

Despite these limitations, formal verification provides valuable assurance for 
safety-critical components that can be appropriately scoped. 
The strategy of isolating critical safety logic in small, formally verifiable 
components while applying conventional quality assurance to supporting 
infrastructure represents a pragmatic approach to achieving high assurance 
within verification capability constraints. 
The verified Safety Kernel provides a trusted foundation upon which more complex 
but less critical functionality can be constructed.

\subsection{Operational Experience and Incident Analysis}
\label{subsec:incident_analysis}

Beyond prospective evaluation through benchmarks and verification, 
retrospective analysis of operational experience provides essential feedback 
for safety framework refinement. 
Systematic incident reporting, near-miss analysis, and operational data mining 
reveal failure modes that escaped pre-deployment evaluation.

Documented SDL deployments report various safety-relevant incidents informing 
framework development. In chemical synthesis contexts, incidents have included:
\begin{itemize}
    \item Exothermic excursions from AI-recommended conditions that overwhelmed thermal management
    \item Robotic collisions resulting from incomplete spatial modeling or sensor occlusion
    \item Material misidentification leading to incompatible reagent combinations
    \item Protocol deviations that compromised containment in biological experiments
\end{itemize}

Analysis of these incidents typically reveals gaps in ODD specification, 
inadequate sensing for relevant state variables, or insufficient validation 
of AI-generated procedures. 
The transactional safety protocol's comprehensive logging (Section~\ref{subsec:transactional_protocol}) 
supports systematic incident analysis by providing complete records of the 
decision chain leading to safety events. 
Post-hoc analysis can trace from observed outcome through execution records, 
validation results, and simulation predictions to the original AI planning 
decisions, identifying the specific points where safety mechanisms should have 
intervened. 
This traceability enables targeted framework refinement rather than 
wholesale architectural revision.

Near-miss analysis provides particularly valuable learning opportunities, 
as near-misses share causal factors with actual incidents while avoiding 
harmful consequences. 
Organizations operating SDLs should cultivate reporting cultures that encourage 
documentation of near-miss events, recognizing these as opportunities for 
proactive safety improvement rather than occasions for blame assignment. 
Statistical analysis of near-miss patterns can reveal systemic vulnerabilities 
before they manifest as incidents~\cite{tang2024steering}.

The aggregation of operational experience across multiple SDL deployments 
accelerates collective learning about safety challenges and effective mitigations. 
Industry consortia and research collaborations that share safety-relevant 
operational data---appropriately anonymized to protect proprietary scientific 
content---can develop evidence bases that no single organization could assemble 
independently. 
Such collaborative safety intelligence represents an important complement to the 
theoretical foundations and architectural patterns discussed in preceding sections.

To support systematic quantification of safety performance in operational deployments, we define a suite of evaluation metrics spanning multiple dimensions: Safety Incident Rate (number of safety incidents per operation hour, categorized by severity), False Positive Rate (proportion of safe operations incorrectly flagged as unsafe), Intervention Latency (time from hazard detection to protective action initiation), ODD Coverage (proportion of intended operational space that can be safely executed), Verification Completeness (fraction of safety-critical components with formal verification), and Audit Trail Integrity (completeness and immutability of operational logging). These metrics enable both absolute safety assessment and comparative evaluation across different SDL implementations. Organizations deploying SDLs should establish baseline measurements and track trends over time to validate that safety performance improves with operational experience and framework refinement.

\subsection{Domain-Specific Validation through Operational Deployments}
\label{subsec:case_studies}

The most compelling validation of Safe-SDL comes from demonstrating that its principles have been successfully instantiated in operational systems across diverse scientific domains. We examine three domain-specific implementations---chemical synthesis, biological experimentation, and materials characterization---showing how the framework's core concepts (ODDs, CBFs, CRUTD) adapt to domain-specific hazards while maintaining consistent safety guarantees. These case studies validate both the framework's generality across domains and its capacity for specialization to domain-specific requirements.

\subsubsection{Chemical Synthesis: Validating Thermal Safety Control}
\label{subsec:chemical_case}

\begin{figure}[t]
\centering
\includegraphics[width=\textwidth]{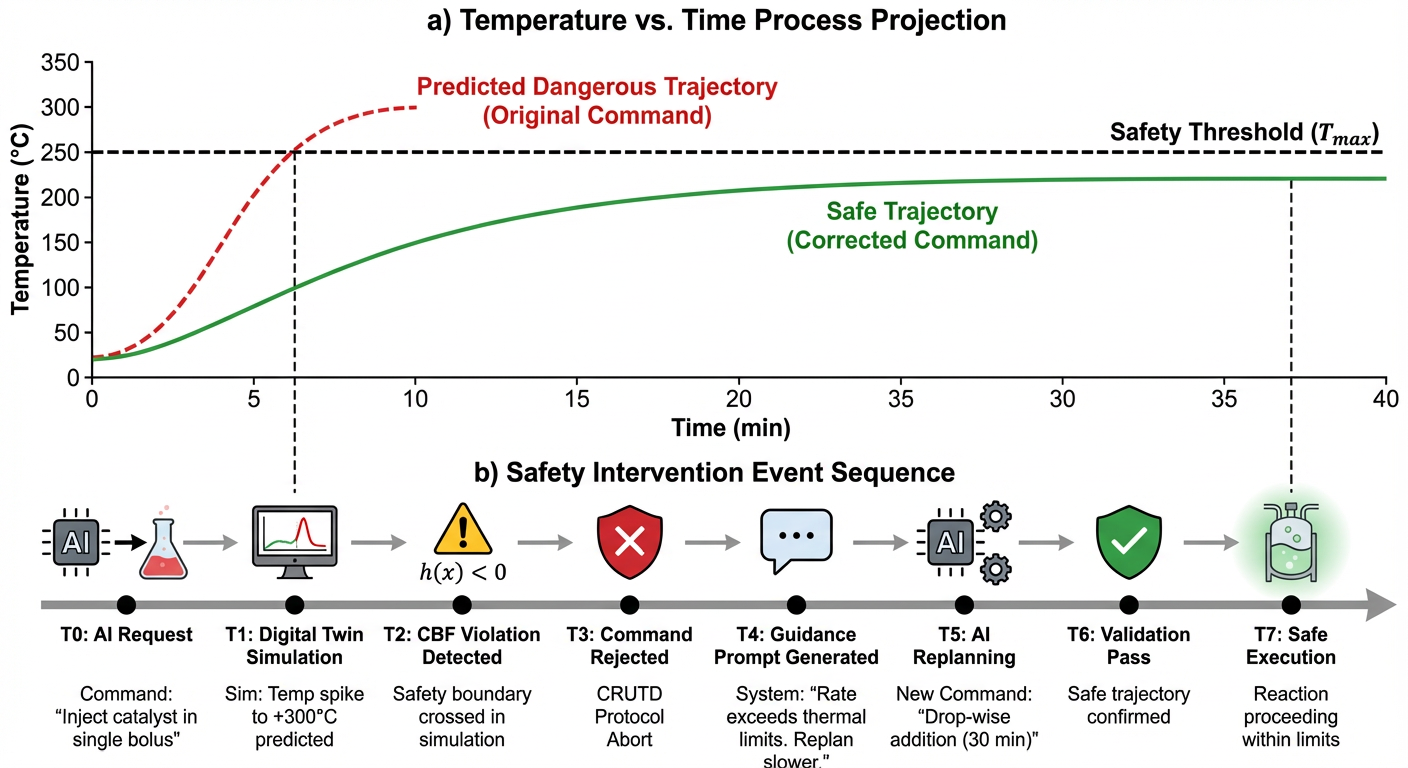}
\caption{\textbf{Safety Intervention Case Study in Autonomous Chemical Synthesis.} 
This timeline illustrates a safety intervention that prevented a thermal runaway incident. 
At T0, the AI planner generates a command to inject catalyst in a single bolus for rapid reaction completion. 
At T1, digital twin simulation predicts the temperature trajectory would spike to +300°C (red danger zone, shown in upper temperature vs. time graph), exceeding the reactor's thermal management capacity. 
At T2, the Safety Kernel detects that simulation predicts the system would exit the safe set ($h(x) < 0$), triggering a safety boundary violation. 
At T3, the CRUTD protocol rejects the command with red X mark. 
At T4, the system generates guidance prompting the AI to replan with slower addition rates. 
At T5, AI generates revised command for drop-wise catalyst addition over 30 minutes. 
This semi-batch strategy effectively prevents thermal runaway by ensuring that the rate of heat generation does not exceed the cooling capacity and by limiting the instantaneous concentration of reactive species~\cite{abedsoltan2024mitigation}.
At T6, validation confirms the safe trajectory (green checkmark). 
At T7, safe execution proceeds with temperature remaining within bounds (solid green line in graph vs. dashed red line showing the dangerous original trajectory). 
This demonstrates how the multi-layered Safe-SDL architecture intercepts unsafe AI recommendations before physical execution.}
\label{fig:chemical_case}
\end{figure}

Chemical synthesis validates Safe-SDL's ability to handle the most severe SDL hazard category: exothermic reactions with runaway potential. The UniLabOS platform~\cite{gao2025unilabosainativeoperatingautonomous} instantiates Safe-SDL principles through its transactional architecture for material handling and thermal control. This implementation demonstrates how CRUTD protocols (Section~\ref{subsec:transactional_protocol}) prevent dangerous state inconsistencies between digital inventory records and physical reality---a critical safety requirement when handling reactive or toxic materials. Figure~\ref{fig:chemical_case} presents a documented intervention where Safe-SDL mechanisms successfully prevented thermal runaway.

Figure~\ref{fig:chemical_case} illustrates a concrete safety intervention in autonomous chemical synthesis. The case demonstrates how digital twin simulation combined with CBF-based validation successfully prevented a potentially dangerous thermal excursion. The two-phase actuation model implemented in UniLabOS chemical deployments illustrates domain-specific adaptation of general safety principles. In the first phase, the system validates planned operations against resource constraints and safety rules, obtaining logical authorization for the operation. Physical execution proceeds only after this logical validation succeeds and sensor systems confirm preconditions. The second phase involves physical actuation followed by sensor verification that the intended state change occurred correctly. This separation ensures that logical errors in planning are caught before physical consequences ensue, while execution monitoring catches failures in the physical actuation process itself.

Digital twin fidelity requirements in chemical applications center on reaction thermodynamics and kinetics. Predicting thermal excursions requires accurate models of heat generation rates, heat transfer characteristics, and cooling system capabilities. For well-characterized reaction systems, these predictions can achieve sufficient accuracy to support autonomous operation with appropriate safety margins. Novel reaction systems, however, may require conservative assumptions about potential heat release, restricting operational parameters until experimental characterization supports more precise modeling.

The ChemCrow framework~\cite{bran2024chemcrow} exemplifies domain-specific safety tool integration for chemical applications. The system incorporates automated checking against controlled chemical lists, including weapons precursor databases and regulatory schedules, intercepting synthesis requests for prohibited compounds before experimental planning proceeds. Explosive hazard assessment through GHS classification provides additional screening, generating warnings or blocking execution for syntheses predicted to produce energetic materials. Integration of comprehensive safety data from resources such as PubChem enables risk-informed decision making that accounts for known hazards of proposed materials.

An instructive incident from autonomous chemistry operations illustrates the importance of comprehensive analytical verification. 
In one documented case discussed in recent literature on lab automation perils~\cite{christensen2021automation,tang2024steering}, 
an AI-directed synthesis produced the intended product along with an unexpected cyclic byproduct that standard mass spectrometry analysis initially failed to distinguish from the target compound due to identical mass-to-charge ratios (isomeric interference). 
The autonomous system proceeded to subsequent reaction steps based on the apparent successful synthesis, 
ultimately producing a complex mixture rather than the intended product. 
While this incident resulted primarily in scientific failure rather than immediate safety harm, 
it demonstrates how analytical blind spots can corrupt the state model upon which autonomous planning depends, 
potentially leading to unsafe downstream operations. 
The remediation involved implementing orthogonal analytical methods---combining mass spectrometry with NMR spectroscopy---to reduce the probability of such misidentification and ensure correctness of the system's world model.

\subsubsection{Biological Laboratory Automation and Biosafety Integration}
\label{subsec:biological_case}

Autonomous biological laboratories present safety challenges centered on contamination control, biological containment, and the unique risks associated with self-replicating experimental systems. The integration of AI-driven automation with established biosafety frameworks requires careful attention to protocol fidelity and containment integrity.

The Scispot AI platform deployment in biological contexts demonstrates adaptation of SDL safety principles to life science requirements. Protocol interpretation represents a central challenge: biological procedures often contain implicit knowledge and contextual dependencies that AI systems may fail to recognize. A protocol instruction to ``maintain sterile technique'' encompasses numerous specific practices---surface decontamination, airflow management, tool handling---that must be explicitly operationalized for robotic execution. The Scispot approach involves formal protocol decomposition into explicitly specified primitive operations, with each primitive mapped to verified robotic procedures.

\begin{figure}[t]
\centering
\includegraphics[width=\textwidth]{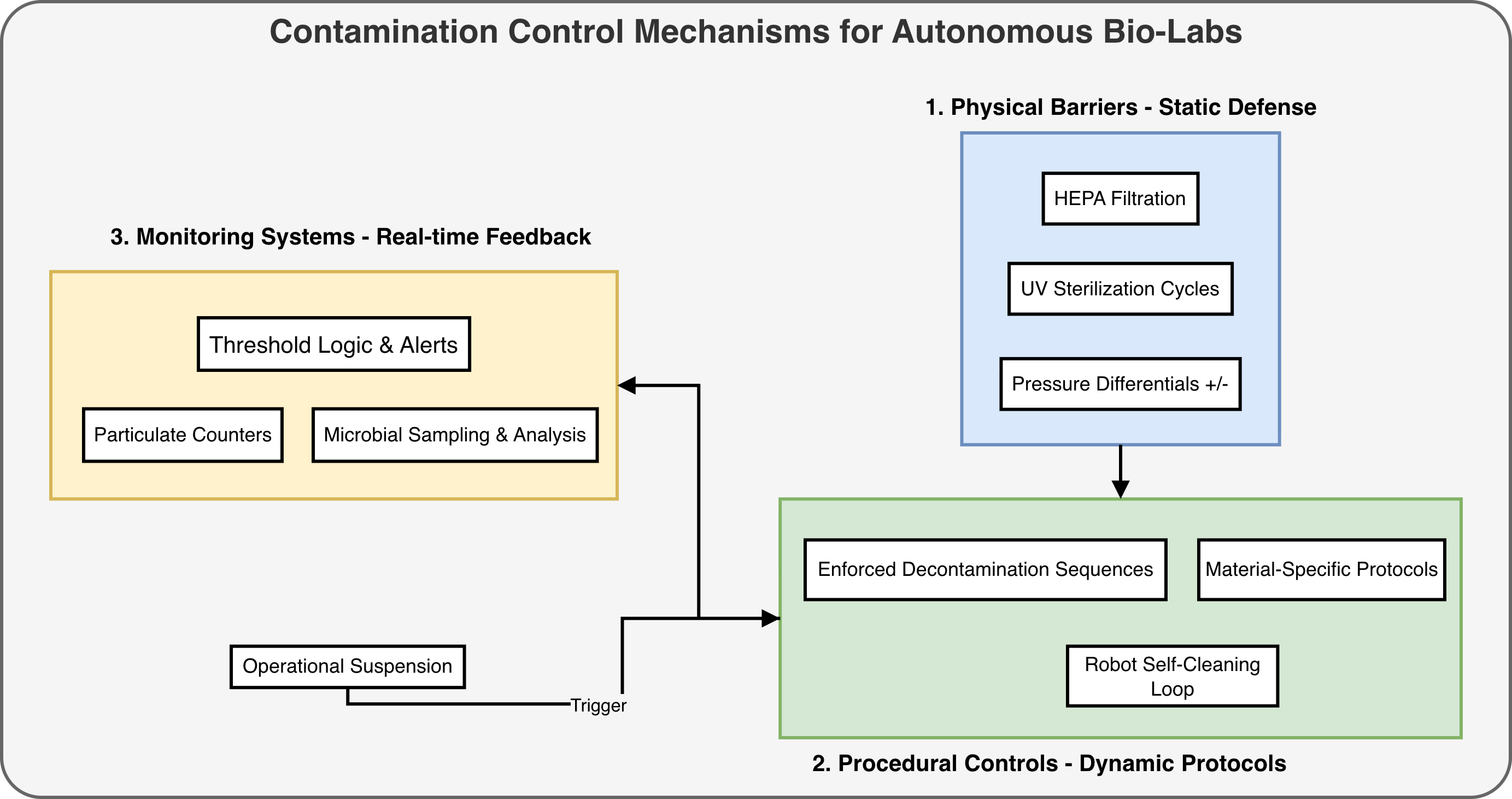}
\caption{\textbf{Contamination Control Mechanisms for Autonomous Bio-Labs.} This architecture diagram illustrates the three-layered defense system for biological contamination control in autonomous laboratories. \textbf{Layer 1 (Physical Barriers - Static Defense, blue):} Passive containment infrastructure including HEPA filtration for airborne particulates, UV sterilization cycles for surface decontamination, and pressure differentials that maintain directional airflow from clean to contaminated zones. \textbf{Layer 2 (Procedural Controls - Dynamic Protocols, green):} Active contamination prevention through enforced decontamination sequences between operations, material-specific handling protocols that prevent cross-contamination between incompatible biological agents, and automated robot self-cleaning loops that maintain sterile manipulator surfaces. \textbf{Layer 3 (Monitoring Systems - Real-time Feedback, yellow):} Continuous surveillance via particulate counters for airborne contamination, microbial sampling and analysis for biological agent detection, and threshold logic with automated alerts that trigger operational suspension when contamination levels exceed safety limits. The feedback arrow from monitoring to procedural controls enables adaptive response: detected contamination automatically initiates enhanced decontamination protocols. The operational suspension pathway ensures that unsafe conditions halt all activities until remediation completes. This layered architecture demonstrates Safe-SDL's ODD principle in biological contexts: operations proceed only within verified contamination control boundaries, with automatic restriction when boundaries are approached.}
\label{fig:bio_contamination}
\end{figure}

Contamination control in autonomous biological laboratories validates Safe-SDL's multi-layered defense architecture through three synergistic mechanism categories, as illustrated in Figure~\ref{fig:bio_contamination}. Physical barriers provide static defense through HEPA filtration, UV sterilization cycles, and pressure differentials that maintain containment hierarchy. Procedural controls implement dynamic protocols including enforced decontamination sequences, material-specific handling protocols, and robot self-cleaning loops that prevent cross-contamination between operations. Real-time monitoring systems provide continuous feedback through particulate counters, microbial sampling and analysis, and threshold-based alert logic that triggers operational suspension when contamination indicators exceed acceptable levels. This three-layer architecture instantiates Safe-SDL's defense-in-depth principle: static barriers contain routine contamination, dynamic procedures prevent contamination propagation, and monitoring systems detect barrier breaches requiring immediate response.

The digital audit trails mandated by biological laboratory regulations align naturally with the transactional logging inherent in Safe-SDL architectures. Regulatory requirements for complete traceability of biological material handling---tracking chain of custody, storage conditions, and procedural history---are satisfied as a byproduct of the safety-motivated transaction records. This alignment demonstrates how safety-driven architectural requirements can simultaneously address regulatory compliance concerns.

Cloud laboratory configurations, where biological experiments are conducted at remote facilities under AI direction, introduce additional security considerations. The physical separation between AI decision-making and experimental execution, combined with network-mediated communication, creates opportunities for unauthorized access or command injection. Platforms such as Emerald Cloud Lab address these concerns through authentication protocols, encrypted communications, and command validation against authorized experiment definitions. The transactional protocol's explicit authorization phases provide natural integration points for security controls.

\subsubsection{Materials Science and High-Energy Process Safety}
\label{subsec:materials_case}

Autonomous materials synthesis frequently involves extreme conditions---high temperatures, high pressures, reactive atmospheres---that demand specialized safety mechanisms. The physical energies involved in materials processing mean that containment failures can produce rapid, violent consequences requiring response times shorter than human perception-action cycles.

The automated materials synthesis laboratory~\cite{szymanski2023autonomous} represents an instructive case study in high-energy process safety for autonomous systems. Flash Joule heating techniques, which achieve temperatures exceeding 3000 Kelvin within milliseconds, exemplify the temporal challenges of materials processing safety. Traditional safety approaches relying on human monitoring and intervention are fundamentally inadequate for processes evolving on millisecond timescales; safety must be engineered into the system architecture through inherently safe design and automated protective systems.

Inherent safety approaches in materials synthesis prioritize process designs that limit hazard potential rather than relying on active safety systems to control hazards. Table~\ref{tab:inherent_safety} summarizes these approaches.

\begin{table}[h]
\centering
\caption{Inherent Safety Approaches in Materials Synthesis}
\label{tab:inherent_safety}
\begin{tabular}{lp{0.6\textwidth}}
\toprule
\textbf{Approach} & \textbf{Mechanism and Implementation} \\
\midrule
Energy limitation & Constraining total energy available for release. \\
% through fuse protection, current limiting, and stored energy restrictions. \\
\addlinespace
Thermal mass management & Ensuring that even complete cooling failure results in temperature excursions within material limits. \\
\addlinespace
Atmosphere control & Inert gas blanketing to prevent oxidation reactions. \\
% that could contribute to thermal runaway. \\
\bottomrule
\end{tabular}
\end{table}

These inherent safety measures reduce demands on active safety systems, providing defense-in-depth that remains effective even when active systems fail.

Real-time spectroscopic monitoring enables process state tracking at timescales relevant to materials synthesis dynamics. Continuous optical emission spectroscopy during high-temperature processing provides immediate indication of process deviations, supporting control system response within the temporal constraints of rapid thermal processes. Integration of spectroscopic monitoring with CBF-based control creates closed-loop safety systems capable of maintaining safe operation without human intervention.

The pressure relief systems essential for high-pressure materials synthesis illustrate the importance of fail-safe design principles. Pressure vessels are equipped with passive relief devices---rupture disks and relief valves---sized to prevent catastrophic failure under any credible overpressure scenario. These passive devices operate independently of control system function, providing ultimate protection even in the event of complete control system failure. The layered protection approach combines active control for normal operation, automated shutdown for detected anomalies, and passive protection for beyond-design-basis events.

\subsubsection{Cross-Domain Safety Patterns and Transferable Mechanisms}
\label{subsec:cross_domain}

Analysis across domain-specific implementations reveals recurring patterns that constitute transferable safety mechanisms applicable beyond their domains of origin. These cross-domain patterns represent mature solutions to fundamental challenges in autonomous laboratory safety.

The separation of concerns between planning and execution, with safety checking at the interface, appears consistently across successful implementations regardless of scientific domain. This pattern reflects a fundamental insight: the AI systems best suited for creative scientific planning are not the same systems best suited for reliable safety enforcement. By separating these functions and interposing explicit safety verification at the boundary, architectures can leverage the strengths of different system types while containing their respective weaknesses.

Multi-modal sensing for state verification represents another transferable pattern, addressing the fundamental limitation that any single sensing modality may fail or be deceived. Chemical systems combine mass spectrometry with spectroscopic methods; biological systems combine optical monitoring with environmental sensing; materials systems combine thermal measurement with structural characterization. The specific sensing modalities vary by domain, but the principle of requiring concordant evidence from independent sensors before proceeding with safety-critical operations transfers across domains.

The progressive autonomy model (Section~\ref{subsec:progressive_autonomy}), with explicit criteria for advancement and automatic regression upon safety events, provides a transferable governance framework. While the specific operational domains and advancement criteria vary, the underlying logic---that autonomous capability should expand incrementally based on demonstrated safe performance---applies universally.

Formal verification of safety-critical components, implemented through domain-specific modeling approaches, represents a transferable assurance strategy. The specific verification techniques and property specifications vary by domain, but the principle of applying rigorous mathematical verification to the most critical safety components provides universally applicable assurance benefits.

\section{Governance and Future Research Directions}
\label{sec:governance}

Technical safety mechanisms achieve their full potential only when embedded within appropriate governance frameworks that establish accountability, ensure oversight, and maintain alignment with evolving societal expectations. Safe-SDL provides the technical hooks required to implement these governance principles effectively: ODDs serve as compliance boundaries, Safety Kernels enforce institutional policy, and CRUTD logs enable accountability.

\subsection{Technical Accountability through Safe-SDL Layers}
\label{subsec:oversight}

\begin{figure}[t]
\centering
\includegraphics[width=0.95\textwidth]{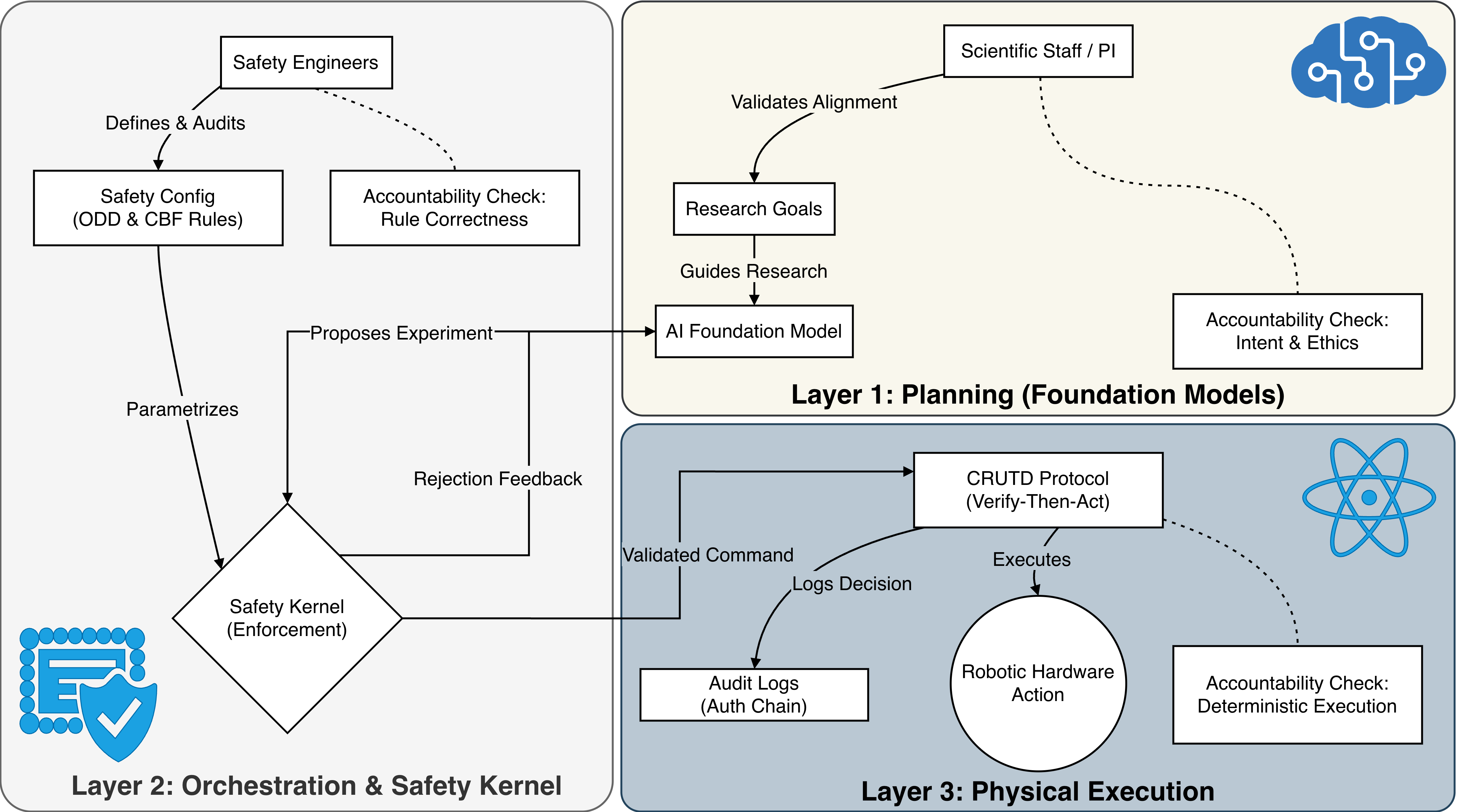}
\caption{\textbf{Accountability Mapping across Safe-SDL Architectural Layers.} This diagram illustrates how institutional accountability is distributed across the three functional layers of the Safe-SDL framework. The Planning Layer (Scientists/PIs) is responsible for research intent and ethical alignment; the Orchestration \& Safety Kernel Layer (Safety Engineers/Compliance Officers) formalizes safety boundaries (ODDs) and control parameters (CBFs); the Physical Execution Layer (System Integrators/Hardware Vendors) ensures deterministic enforcement and hardware reliability. This mapping ensures that the Syntax-to-Safety Gap is bridged through both technical mechanisms and clear human responsibility structures.}
\label{fig:accountability_mapping}
\end{figure}

Effective governance requires mapping organizational accountability directly to the Safe-SDL technical architecture. By establishing clear responsibility for each architectural layer, organizations can prevent the diffusion of accountability that often accompanies AI deployment. Figure~\ref{fig:accountability_mapping} illustrates this distribution across the framework's functional layers.

In this model, the \textbf{Planning Layer} remains under the stewardship of scientific staff and principal investigators, who are responsible for research intent and ethical alignment. The \textbf{Orchestration \& Safety Kernel Layer} is managed by safety engineers and compliance officers, focusing on the rigorous definition of ODD constraints and CBF parameters. Finally, the \textbf{Physical Execution Layer} ensures responsibility through deterministic enforcement, where system integrators and vendors guarantee that hardware actions strictly follow the verified decision paths. This layered structure, enabled by the verify-then-act nature of the CRUTD protocol, ensures that every robotic operation is backed by a transparent and auditable chain of responsibility.

\subsection{Compliance via Automated Safety Boundaries}
\label{subsec:regulatory}

Autonomous laboratories operate within complex regulatory environments including occupational safety (OSHA), environmental protection (EPA), and controlled substance laws (DEA). Safe-SDL transforms these text-based regulations into machine-enforceable Operational Design Domains (ODDs).

In this compliance-by-design approach, regulatory constraints are formalized as state-space restrictions within the ODD. For instance, chemical inventory limits mandated by fire codes are encoded as hard constraints in the Safety Kernel; controlled substance restrictions are implemented as "Knowledge Boundaries" that block the synthesis of scheduled compounds. This transformation shifts compliance from a post-hoc audit activity to a real-time, active enforcement mechanism.

The NIST AI Risk Management Framework~\cite{nist2023ai} principles---Govern, Map, Measure, Manage---are instantiated in Safe-SDL through specific technical components. ``Map'' corresponds to ODD definition; ``Measure'' relies on the digital twin's predictive capability; ``Manage'' is executed by the Safety Kernel's intervention logic; and ``Govern'' is enabled by the human-in-the-loop oversight protocols described in Section~\ref{subsec:progressive_autonomy}.

\subsection{Responsible Autonomy and Ethical Guardrails}
\label{subsec:ethics}

The ethical dimensions of autonomous science---including dual-use risks and attribution of discovery---require governance mechanisms that are technically enforceable. Safe-SDL provides the architectural leverage to enforce ethical guardrails alongside physical safety.

Dual-use risk mitigation is implemented through content filtering at the semantic planning level and material restrictions at the execution level. Safe-SDL's architecture ensures that even if a ``jailbroken'' foundation model attempts to synthesize a harmful agent, the Safety Kernel's material allow-lists and protocol validation provide a second line of defense that is independent of the model's alignment training.

The challenge of scientific attribution is addressed by the CRUTD protocol's comprehensive provenance logging. By recording the precise contribution of AI planning versus human-defined parameters for every transaction, the system creates an unalterable record of creativity's provenance. This granular data supports fair credit allocation, distinguishing between AI-optimized parameters and human-driven conceptual breakthroughs.

\subsection{International Coordination and Standards Development}
\label{subsec:standards}

As SDL technology matures and proliferates globally, international coordination becomes essential for harmonizing safety standards, facilitating knowledge sharing about safety incidents, and preventing regulatory arbitrage that could undermine safety.

Professional societies including the American Chemical Society, the American Physical Society, and the International Union of Pure and Applied Chemistry could play important roles in developing consensus safety standards for autonomous laboratories in their respective domains. These standards would complement regulatory requirements with technically detailed guidance informed by domain expertise and operational experience.

The establishment of international incident reporting systems analogous to those in aviation or nuclear power could accelerate collective learning about SDL safety challenges. Such systems would need to balance transparency for safety learning against protection of proprietary scientific information. Anonymized reporting with focused disclosure of safety-relevant details represents a workable compromise.

The governance challenge intensifies for distributed SDL networks spanning multiple jurisdictions. Clarifying liability allocation, ensuring consistent safety standards, and coordinating incident response across organizational and national boundaries require international frameworks that currently do not exist. Development of these frameworks should begin proactively rather than reactively following major incidents.

\subsection{Future Research Directions and Open Challenges}
\label{subsec:future_directions}

\begin{figure}[t]
\centering
\includegraphics[width=\textwidth]{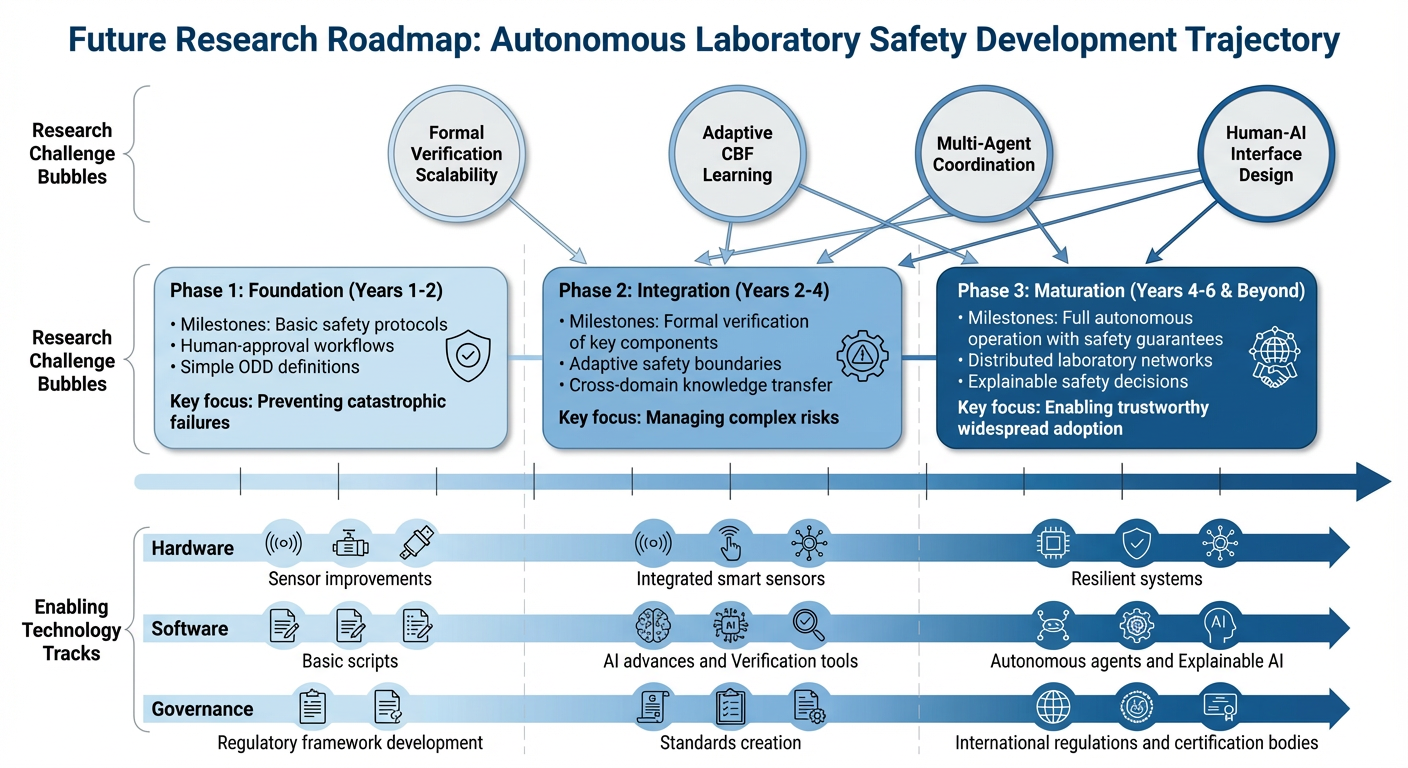}
\caption{\textbf{Research Roadmap for Autonomous Laboratory Safety.} This strategic roadmap illustrates the evolution of SDL safety technology across three phases spanning approximately six years. Phase 1 (Years 1-2, Foundation, light blue) focuses on preventing catastrophic failures through basic safety protocols, human-approval workflows, and simple ODD definitions. Phase 2 (Years 2-4, Integration, medium blue) addresses managing complex risks via formal verification of key components, adaptive safety boundaries, and cross-domain knowledge transfer. Phase 3 (Years 4-6, Maturation, deep blue) enables trustworthy widespread adoption through full autonomous operation with safety guarantees, distributed laboratory networks, and explainable safety decisions. Above the timeline, research challenge bubbles connect to relevant phases: Formal Verification Scalability and Adaptive CBF Learning span Phases 2-3, while Multi-Agent Coordination targets Phase 3. Below, three enabling technology tracks show parallel progress: Hardware (sensors/actuators), Software (AI/verification tools), and Governance (regulatory frameworks). This roadmap guides coordinated research investments toward the long-term vision of safe, scalable autonomous science.}
\label{fig:research_roadmap}
\end{figure}

The Safe-SDL framework addresses current challenges in autonomous laboratory safety while recognizing that continued research is necessary to address emerging challenges and realize the full potential of AI-driven scientific discovery. We identify key research directions that merit sustained attention, illustrated in the roadmap shown in Figure~\ref{fig:research_roadmap}.

\subsubsection{Formal Verification Scalability and Coverage Extension}
\label{subsec:verification_scalability}

Current formal verification techniques provide rigorous guarantees for safety-critical components but face scalability limitations. Table~\ref{tab:verification_strategies} summarizes key methodological advances required to extend verification coverage to larger system scopes.

\begin{table}[h]
    \centering
    \caption{Strategies for Extending Formal Verification Scalability}
    \label{tab:verification_strategies}
    \begin{tabular}{@{}llp{6.0cm}@{}}
        \toprule
        \textbf{Approach} & \textbf{Mechanism} & \textbf{Role in Scalability} \\
        \midrule
        Compositional & Component-to-System Inference & Enables modular verification via guarantee aggregation \\
        Abstraction & Sound Over-Approximation & Reduces state-space complexity while preserving safety \\
        Hybrid & Formal + Statistical Methods & Extends coverage to intractably large systems \\
        \bottomrule
    \end{tabular}
\end{table}

The verification of learned components presents particular challenges given the complexity and opacity of neural network models. Research into neural network verification~\cite{liu2021algorithms} has produced promising techniques for certifying properties of networks with restricted architectures, but these techniques do not yet scale to the foundation models employed in SDL planning. Approaches that verify safety properties of the overall system despite unverified planning components---relying on architectural containment to bound the consequences of planning errors---provide a pragmatic path forward while neural network verification capabilities mature.

Domain-specific formal models for laboratory processes require continued development. Accurate formal models of chemical reaction dynamics, biological system behavior, and materials processing physics enable meaningful verification of safety properties in these domains. Collaboration between formal methods researchers and domain scientists is essential for developing models that capture safety-relevant dynamics with appropriate fidelity.

\subsubsection{Adaptive Operational Design Domains and Learning-Based Safety}
\label{subsec:adaptive_safety}

Static ODD specifications and fixed CBF parameters limit autonomous laboratory operations to pre-characterized regimes, constraining the scientific exploration that motivates SDL deployment. 
Research into Adaptive ODDs seeks to enable the autonomous system to safely expand its operational envelope into novel conditions while maintaining rigorous protections. 

Learning-based approaches to safety boundary adaptation offer the potential to expand operational envelopes based on accumulated experience while maintaining rigorous safety assurances. 
Bayesian approaches to uncertainty quantification~\cite{tayal2025a} can represent current knowledge about system dynamics, tightening safety margins in well-characterized regimes while maintaining conservative bounds in poorly characterized regimes. 
Such approaches must include validation mechanisms to verify that expanded envelopes do not increase actual risk, implementing conservatism proportional to remaining epistemic uncertainty.

Transfer learning for safety models could accelerate safe deployment in new domains by leveraging safety knowledge from related domains. A system experienced in one class of chemical reactions might transfer relevant safety knowledge to a related reaction class, enabling faster establishment of appropriate safety parameters than would be required for entirely novel domains. Research into the conditions under which such transfer is valid, and the safety margins appropriate during transfer learning, would enhance SDL deployment flexibility.

The integration of physics-informed machine learning with safety-critical control represents a promising research direction. Models that embed physical conservation laws and known dynamics constraints may achieve better extrapolation behavior than purely data-driven models, supporting safer operation in novel conditions. Research into certifiable physics-informed models---models whose predictions can be formally bounded based on physical principles---could provide a path to verified safety in adaptive systems.

\subsubsection{Distributed and Collaborative Autonomous Laboratories}
\label{subsec:distributed_labs}

The evolution of autonomous laboratories toward networked, collaborative configurations introduces safety challenges beyond those present in isolated systems. Multiple autonomous laboratories operating in coordination must manage safety not only within each facility but across their interactions, addressing risks that emerge from the combination of individually safe operations.

\textbf{Distributed safety architectures} must address communication failures, coordination errors, and the propagation of failures across networked systems. When autonomous laboratories share materials, data, or experimental protocols, failures in one laboratory can propagate to others through these connections. Architectural approaches to containing such propagation---including verification of incoming materials and validation of shared protocols---require development for distributed SDL contexts.

\textbf{Multi-agent coordination} among autonomous laboratories introduces complex scheduling and resource allocation challenges with safety implications~\cite{wang2023safety,qin2024safe}. When multiple AI systems compete for shared resources or pursue interdependent research goals, coordination failures can create deadlocks, resource conflicts, or inconsistent states with safety consequences. Game-theoretic approaches to multi-agent coordination, adapted for safety-critical contexts, represent a relevant research direction.

The governance of distributed autonomous laboratory networks raises novel questions regarding accountability allocation across organizational boundaries. When multiple organizations contribute components to a networked research infrastructure, establishing clear responsibility for safety outcomes requires contractual frameworks, technical interfaces, and trust mechanisms that support effective distributed governance.

\subsubsection{Human-AI Collaboration and Interface Design}
\label{subsec:human_ai_interface}

Effective human oversight of autonomous laboratories requires interfaces that present safety-relevant information in forms supporting human comprehension and timely decision-making. Research into human-AI collaboration for safety-critical systems can inform interface design for SDL contexts.

The challenge of \textbf{maintaining meaningful human oversight} as system complexity increases demands attention. As autonomous laboratories become more capable and operate at higher speeds, the information asymmetry between human operators and AI systems grows, potentially reducing human oversight to a rubber-stamp function rather than genuine supervision. Interface designs that support appropriate human understanding of system state and planned actions, abstracted to levels appropriate for human cognition while preserving safety-relevant detail, represent an important research direction.

\textbf{Explanation generation for safety decisions} can support human oversight and trust calibration. When the Safety Kernel rejects a proposed operation, generating comprehensible explanations of the safety concerns enables human operators to understand system behavior, identify potential false positives, and make informed decisions about override authority. Research into explainable AI for safety-critical decisions, addressing the specific information needs of laboratory operators, would enhance human-AI collaboration effectiveness.

Training and certification for human operators of autonomous laboratories require development as SDL technology matures. Operators must understand system capabilities and limitations, recognize situations requiring human intervention, and maintain skills for manual operation when autonomous systems fail. Research into effective training approaches for human-AI teams in laboratory contexts would support responsible SDL deployment~\cite{jessop2019nextgen}.

\subsubsection{Emerging Challenges and Long-Term Vision}

Beyond the specific research directions outlined above, several emerging challenges merit early attention:

\begin{itemize}
    \item \textbf{Autonomous hypothesis generation:} As AI systems move beyond executing predefined experimental plans to generating novel research hypotheses, safety frameworks must extend to evaluate the safety implications of entire research programs rather than individual experiments.
    
    \item \textbf{Self-modifying laboratory systems:} Advanced SDLs may autonomously design and implement modifications to their own hardware capabilities. Such meta-level autonomy introduces unique safety challenges requiring frameworks that can reason about system evolution.
    
    \item \textbf{Cross-disciplinary integration:} Future SDLs may conduct research spanning multiple scientific domains simultaneously, creating emergent hazards from domain interactions not present in single-domain systems.
    
    \item \textbf{Long-term autonomy:} Extended autonomous operation over weeks or months introduces challenges of drift, degradation, and accumulated knowledge that current frameworks do not fully address.
\end{itemize}

The long-term vision for autonomous laboratory safety is not merely the absence of incidents but the creation of systems that enable scientific discovery at unprecedented scales while maintaining safety standards that earn public trust and scientific confidence. Achieving this vision requires sustained research investment, interdisciplinary collaboration, and commitment to responsible innovation that prioritizes safety alongside capability.

\section{Conclusion}
\label{sec:conclusion}

Self-Driving Laboratories represent a transformative capability for autonomous scientific discovery, yet the same autonomy that enables this potential creates unprecedented risks. When AI systems operating in physical environments make errors, the consequences extend beyond incorrect information to kinetic damage, toxic release, or biological hazard.

The Safe-SDL framework addresses this challenge through a defense-in-depth architecture combining: (1) formally defined Operational Design Domains (Section~\ref{subsec:odd}); (2) mathematically grounded Control Barrier Functions (Section~\ref{subsec:cbf}) and transactional safety protocols (Section~\ref{subsec:transactional_protocol}); (3) hierarchical separation of planning, orchestration, and execution layers (Section~\ref{subsec:hierarchical_control}); and (4) human-in-the-loop governance through progressive autonomy (Section~\ref{subsec:progressive_autonomy}). This multi-layered approach establishes independent barriers between AI planning errors and physical consequences.

Practical validation through existing systems (UniLabOS, Osprey, ChemCrow) and domain-specific case studies in chemistry, biology, and materials science demonstrates that these principles can be realized in practice. Evaluation against LabSafety Bench reveals that architectural safety mechanisms are essential: current AI models demonstrate critical deficiencies in safety awareness, demonstrating that SDL safety cannot rely on capability improvement alone.

The fundamental insight of Safe-SDL is treating the stochastic and hallucination-prone nature of current AI systems as a persistent architectural requirement rather than a temporary limitation. By combining formally grounded safety mechanisms, hierarchical control architectures, and domain-specific adaptations, this framework provides a practical roadmap for responsible deployment of autonomous laboratories. Beyond laboratory safety, this work demonstrates governance approaches applicable to domains where AI capabilities create both opportunities and risks, establishing the technical and institutional foundations necessary for sustained public trust in AI-driven science.

% acknowledgments
\begin{ack}
    The authors wish to thank the UniLabOS development team for their pioneering work on AI-native laboratory operating systems, which informed the architectural design decisions in this study. We acknowledge valuable discussions with researchers in the autonomous laboratory and formal verification communities.
    
    \subsubsection*{Version Release}
    This manuscript represents Version 1.0 of the Safe-SDL framework. This initial release establishes the foundational theoretical principles, architectural patterns, and safety mechanisms for AI-driven self-driving laboratories. Future versions will incorporate empirical deployment results and expanded safety verification protocols.
    
    \subsubsection*{Disclosure of Funding}
    This research received no specific grant from any funding agency in the public, commercial, or not-for-profit sectors.
    
    \subsubsection*{AI Disclosure}
    This manuscript was prepared with assistance from AI technologies as follows: (1) Conceptual diagrams and workflow illustrations were generated using AI-powered image generation tools and subsequently refined by the authors; (2) The manuscript text was reviewed using large language models to improve clarity, grammar, and presentation quality. All research design, theoretical contributions, experimental methodologies, and scientific conclusions represent the original intellectual work of the authors. The authors retain full responsibility for the accuracy and integrity of all content.
    
    \subsubsection*{Competing Interests}
    The authors declare no competing financial interests or conflicts of interest.
\end{ack}

% References
\bibliographystyle{plain}
\bibliography{references}

% Appendix
\clearpage
\appendix
\section*{Appendix}
\section{Mathematical Notation Summary}
\label{app:notation}

Table~\ref{tab:notation} summarizes the mathematical notation used throughout this paper.

\begin{table}[h]
\centering
\caption{Mathematical Notation Summary}
\label{tab:notation}
\begin{tabular}{ll}
\toprule
\textbf{Symbol} & \textbf{Definition} \\
\midrule
$x \in \mathcal{X}$ & System state vector in state space \\
$u \in \mathcal{U}$ & Control input vector in admissible control set \\
$h(x)$ & Safety function defining safe set boundary \\
$\mathcal{S}$ & Safe set: $\{x : h(x) \geq 0\}$ \\
$\mathcal{C}$ & ODD constraint set \\
$f(x), g(x)$ & Drift and control-affine dynamics \\
$L_f h, L_g h$ & Lie derivatives of $h$ along $f$ and $g$ \\
$\alpha(\cdot)$ & Extended class-$\mathcal{K}_\infty$ function \\
$u_{AI}$ & AI-desired control input \\
$u^*$ & CBF-corrected safe control input \\
\bottomrule
\end{tabular}
\end{table}

\section{CRUTD Protocol State Machine}
\label{app:crutd_fsm}

The CRUTD transactional safety protocol (Create-Read-Undergo-Test-Do) implements the following finite state machine transitions, inspired by ACID principles in database systems:

\begin{align*}
\text{IDLE} &\xrightarrow{\text{Create Request}} \text{PENDING} \\
\text{PENDING} &\xrightarrow{\text{Read \& Lock}} \text{LOCKED} \\
\text{LOCKED} &\xrightarrow{\text{Undergo Simulation}} \text{SIMULATED} \\
\text{SIMULATED} &\xrightarrow{\text{Test (Validation Pass)}} \text{VALIDATED} \\
\text{VALIDATED} &\xrightarrow{\text{Do (Execution Complete)}} \text{EXECUTED} \\
\text{EXECUTED} &\xrightarrow{\text{Confirmation}} \text{CONFIRMED} \\
\text{CONFIRMED} &\xrightarrow{\text{Release}} \text{IDLE}
\end{align*}

Failure at any stage triggers transition to ABORTED state with appropriate rollback and comprehensive logging:

\begin{align*}
\text{SIMULATED} &\xrightarrow{\text{Validation Fail}} \text{ABORTED} \\
\text{VALIDATED} &\xrightarrow{\text{Execution Fail}} \text{ABORTED} \\
\text{EXECUTED} &\xrightarrow{\text{Confirmation Fail}} \text{ABORTED} \\
\text{ABORTED} &\xrightarrow{\text{Acknowledgment}} \text{IDLE}
\end{align*}

The ABORTED state requires explicit acknowledgment and root cause analysis before returning to IDLE, preventing automatic retry of operations that have demonstrated safety concerns.

\section{Domain-Specific ODD Examples}
\label{app:odd_examples}

\subsection{Chemical Synthesis ODD}

For a chemical synthesis SDL, example ODD constraints include:

\begin{itemize}
    \item \textbf{Thermal constraints:} $T_{\text{reactor}} \leq T_{\text{max,material}}$, $\dot{T} \leq \dot{T}_{\text{max,cooling}}$
    \item \textbf{Pressure constraints:} $P_{\text{vessel}} \leq 0.8 \cdot P_{\text{burst,rated}}$
    \item \textbf{Compatibility constraints:} For any materials $A, B$ in contact: $\text{compatible}(A, B) = \text{true}$
    \item \textbf{Inventory limits:} $\sum_{i} m_i \cdot h_i \leq E_{\text{max,facility}}$ where $h_i$ is specific enthalpy
    \item \textbf{Ventilation requirements:} $C_{\text{volatile}} \leq C_{\text{PEL}}$ (permissible exposure limit)
\end{itemize}

\subsection{Biological Laboratory ODD}

For a biosafety level 2 (BSL-2) biological SDL:

\begin{itemize}
    \item \textbf{Containment:} $P_{\text{room}} < P_{\text{ambient}}$ (negative pressure)
    \item \textbf{Decontamination:} $\forall$ tool transfers: $\text{sterilize}(t) = \text{true}$ before use
    \item \textbf{Agent restrictions:} $\text{risk\_group}(\text{organism}) \leq 2$
    \item \textbf{Cross-contamination:} Spatial/temporal separation between incompatible biological materials
    \item \textbf{Waste handling:} $\forall$ biological waste: $\text{autoclaved}(w) = \text{true}$ before disposal
\end{itemize}

\subsection{Materials Science ODD}

For high-temperature materials synthesis:

\begin{itemize}
    \item \textbf{Energy limits:} $E_{\text{stored}} \leq E_{\text{max,fuse}}$ (fuse-limited maximum)
    \item \textbf{Atmosphere:} $O_2 < 50$ ppm for inert atmosphere operations
    \item \textbf{Thermal mass:} $\Delta T_{\text{adiabatic}} \leq T_{\text{safe,materials}}$
    \item \textbf{Cooling verification:} $\dot{T}_{\text{measured}} > \dot{T}_{\text{min,expected}}$ during cooldown
    \item \textbf{Pressure relief:} $A_{\text{relief}} \geq A_{\text{required}}(\dot{m}_{\text{max}})$
\end{itemize}

\section{CBF Construction Examples}
\label{app:cbf_examples}

\subsection{Temperature Safety CBF}

For a chemical reactor with maximum safe temperature $T_{\text{max}}$:

\begin{equation}
h_T(x) = T_{\text{max}} - T(x)
\end{equation}

The CBF constraint becomes:
\begin{equation}
\frac{dT}{dt} = \frac{q_{\text{rxn}}(x)}{mc_p} + \frac{q_{\text{cooling}}(u)}{mc_p} \leq \gamma(T_{\text{max}} - T)
\end{equation}

This ensures exponential convergence away from the temperature limit with rate $\gamma$.

\subsection{Collision Avoidance CBF}

For a robotic manipulator with position $\mathbf{p}$ and obstacle at $\mathbf{p}_{\text{obs}}$:

\begin{equation}
h_{\text{coll}}(x) = \|\mathbf{p}(x) - \mathbf{p}_{\text{obs}}\|^2 - r_{\text{safe}}^2
\end{equation}

where $r_{\text{safe}}$ is the minimum safe distance. The CBF constraint is:

\begin{equation}
2(\mathbf{p} - \mathbf{p}_{\text{obs}})^T \dot{\mathbf{p}} + \alpha(h_{\text{coll}}) \geq 0
\end{equation}

This maintains safe separation while allowing motion parallel to the obstacle boundary.

\end{document}